# Computational Logic Foundations of KGP Agents

**Antonis Kakas**                                    ANTONIS@UCY.AC.CY
*Department of Computer Science, University of Cyprus*
*75 Kallipoleos Str., P.O. Box 537, CY-1678 Nicosia, Cyprus*

**Paolo Mancarella**                                 PAOLO.MANCARELLA@UNIPI.IT
*Dipartimento di Informatica, Università di Pisa*
*Largo B. Pontecorvo, 3 - 56127 Pisa, Italy*

**Fariba Sadri**                                     FS@DOC.IC.AC.UK
*Department of Computing, Imperial College London*
*South Kensington Campus, London SW72AZ, UK*

**Kostas Stathis**                                   KOSTAS@CS.RHUL.AC.UK
*Department of Computer Science, Royal Holloway*
*University of London, Egham, Surrey TW20 0EX, UK*

**Francesca Toni**                                   FT@DOC.IC.AC.UK
*Department of Computing, Imperial College London*
*South Kensington Campus, London SW72AZ, UK*

## Abstract

This paper presents the computational logic foundations of a model of agency called the *KGP* (Knowledge, Goals and Plan) model. This model allows the specification of heterogeneous agents that can interact with each other, and can exhibit both proactive and reactive behaviour allowing them to function in dynamic environments by adjusting their goals and plans when changes happen in such environments. KGP provides a highly modular agent architecture that integrates a collection of reasoning and physical capabilities, synthesised within transitions that update the agent's state in response to reasoning, sensing and acting. Transitions are orchestrated by cycle theories that specify the order in which transitions are executed while taking into account the dynamic context and agent preferences, as well as selection operators for providing inputs to transitions.

## 1. Introduction

It is widely acknowledged that the concept of *agency* provides a convenient and powerful abstraction to describe complex software entities acting with a certain degree of autonomy to accomplish tasks, often on behalf of a user (Wooldridge, 2002). An agent in this context is understood as a software component with capabilities such as reacting, planning and (inter) acting to achieve its goals in the environment in which it is situated. In this paper, we present a model of agency, called *KGP* (Knowledge, Goals and Plan). The model is hierarchical and highly modular, allowing independent specifications of a collection of reasoning and physical capabilities, used to equip an agent with intelligent decision making and adaptive behaviour. The model is particularly suited to open, dynamic environments where the agents have to adapt to changes in their environment and they have to function in circumstances where information is incomplete.

The development of the KGP model was originally motivated by the existing gap between modal logic specifications (Rao & Georgeff, 1991) of BDI agents (Bratman, Israel, & Pollack, 1988) and their implementation (for example see the issues raised by Rao, 1996). Another motivation for the development of KGP comes from our participation in the SOCS project (SOCS, 2007), where we had the need for an agent model that satisfies several requirements. More specifically, we aimed at an agent model that was rich enough to allow intelligent, adaptive and heterogeneous behaviour, formal so that it could lent itself well to formal analysis, and implementable in such a way that the implementation was sufficiently close to the formal specification to allow verification. Although several models of agency have been proposed, none satisfies all of the above requirements at once.

To bridge the gap between specification and implementation the KGP model is based on computational logic (CL). The focus of the work is to extend and synthesise a number of useful computational logic techniques to produce formal and executable specifications of agents. For this purpose, the model integrates abductive logic programming (ALP) (Kakas, Kowalski, & Toni, 1992), logic programming with priorities (Kakas, Mancarella, & Dung, 1994; Prakken & Sartor, 1997) and constraint logic programming (Jaffar & Maher, 1994). Each of these techniques has been explored in its own right, but their modular integration within the KGP model explores extensions of each, as well as providing the high level agent reasoning capabilities.

The KGP model provides a hierarchical architecture for agents. It specifies a collection of modular knowledge bases, each formalised in CL. These knowledge bases support a collection of *reasoning capabilities*, such as planning, reactivity, and goal decision, all of which are given formal specifications. The model also includes a specification of *physical capabilities*, comprising of sensing and actuating. The capabilities are utilised within *transitions*, that model how the *state* of the agent changes as a result of its reasoning, sensing and acting. Transitions use *selection operators* providing them with inputs. A control component, called *cycle theory*, also formalised in CL, specifies in what order the transitions are executed, depending on the environment, the state of the agent, and the preferences of the agent. The cycle theory takes the agent control beyond the *one-size-fits-all* approach used by most agent models, and allows us to specify agents with different preferences and profiles of behaviour (Sadri & Toni, 2005). In particular, whereas the majority of existing agent models rely upon an "observe-plan-act", by means of our cycle theory we can model behaviours such as "observe-revise goals-plan–act" or "observe-plan-sense action preconditions-act" or "observe-plan-act-plan-act". We provide one example of cycle theory, that we refer to as *normal*, allowing all behaviours above depending on different circumstances (the environment in which the agent is situated and its preferences). Note also that, with respect to other agent models, the KGP model allows agents to revise their goals during their life-time, and observing the environment according to two modalities: active and passive observation.

An agent built with a KGP architecture dynamically determines its goals, plans (partially) how to achieve the goals, interleaves planning with action executions and with making observations in the environment and receiving any messages from other agents, adapts its goals and plans to any new information it receives, and any changes it observes, and generates appropriate reactions.

A number of publications have already described aspects of (an initial version of) the KGP agents. A precursor of the overall model has been described by Kakas, Mancarella,

Sadri, Stathis, and Toni (2004b), its planning component has been presented by Mancarella, Sadri, Terreni, and Toni (2004), its cycle theory has been developed by Kakas, Mancarella, Sadri, Stathis, and Toni (2004a) and its implementation has been discussed by Stathis et al. (2004), by Yip, Forth, Stathis, and Kakas (2005), and by Bracciali, Endriss, Demetriou, Kakas, Lu, and Stathis (2006). In this paper, we provide the full formal specification of all the components of the KGP model, thus offering the complete technical account of KGP in one place. In providing this full formal specification, we have adjusted and further developed the model. In particular, the notion of state and its definition is novel, the reasoning capabilities have been simplified and some have been added, the physical capabilities have been extended (to include actuating) and formally defined, the transitions and the selection operators have been formally defined in full.

The rest of the paper is structured as follows. In Sections 2 and 3 we give an outline of the model and then review the background information necessary for the full description. In Sections 4, 5, 6 and 7, respectively, we describe the internal state of KGP agents, their reasoning and physical capabilities, and their transitions. In Section 8 we describe the selection operators which are then used in the cycle theory which is described in Section 9. Following the detailed description of KGP agents we illustrate the model by a series of examples in Section 10, and then compare the model with others in the literature in Section 11. Finally, we conclude the paper in Section 12.

## 2. KGP Model: Outline

In this Section we give an overview of the KGP agent model and its components, and provide some informal examples of its functioning. This model relies upon

- an internal (or mental) *state*, holding the agent *K*nowledge base (beliefs), *G*oals (desires) and *P*lans (intentions),

- a set of *reasoning capabilities*,

- a set of *physical capabilities*,

- a set of *transition rules*, defining how the state of the agent changes, and defined in terms of the above capabilities,

- a set of *selection operators*, to enable and provide appropriate inputs to the transitions,

- a *cycle theory*, providing the control for deciding which transitions should be applied when.

The model is defined in a modular fashion, in that different activities are encapsulated within different capabilities and transitions, and the control is a separate module. The model also has a hierarchical structure, depicted in Figure 1.

### 2.1 Internal State

This is a tuple $\langle KB_0, \mathcal{F}, \mathcal{C}, \Sigma \rangle$, where:

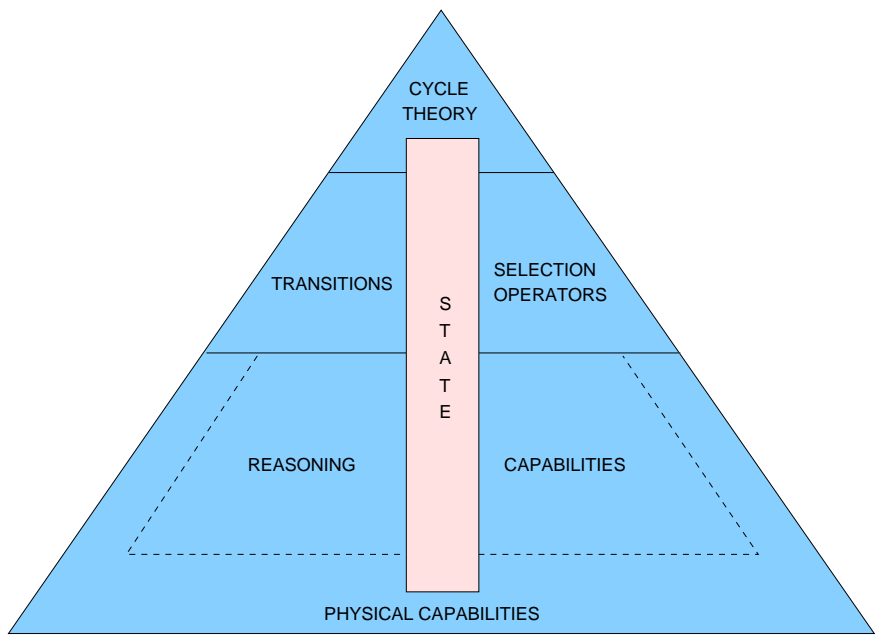

Figure 1: A graphical overview of the KGP model

- $KB_0$ holds the beliefs of the agent about the external world in which it is situated (including past communications), as well as a record of the actions it has already executed.

- $\mathcal{F}$ is a forest of trees whose nodes are *goals*, which may be executable or not. Each tree in the forest gives a hierarchical presentation of goals, in that the tree represents the construction of a plan for the root of the tree. The set of leaves of any tree in $\mathcal{F}$ forms a currently chosen plan for achieving the root of the tree. *Executable goals* are *actions* which may be *physical*, *communicative*, or *sensing*. For simplicity, we assume that actions are atomic and do not have a duration. *Non-executable goals* may be *mental* or *sensing*. Only non-executable mental goals may have children, forming (partial) plans for them. Actions have no children in any tree in $\mathcal{F}$. Each goal has an associated *time variable*, which is implicitly *existentially quantified* within the overall state and serves two purposes: (1) indicating the time the goal is to be achieved, which is instantiated if the goal is achieved at an appropriate time, and (2) providing a unique identifier for that goal. In the remainder of the paper, we will often use the following terminology for goals in $\mathcal{F}$, when we want to emphasise their role and/or their nature: the roots of trees in $\mathcal{F}$ will be referred to as *top-level goals*, executable goals will be referred to as *actions*, and non-executable goals which are not top-level goals will be referred to as *sub-goals*. Top-level goals will be classified as *reactive* or *non-reactive*, as will be explained later. [1] Note that some top-level (reactive) goals may be actions.

---

1. Roughly speaking, reactive goals are generated in response to observations, e.g. communications received from other agents and changes in the environment, for example to repair plans that have already been generated. Non-reactive goals, on the other hand, are the chosen desires of the agent.

- $\mathcal{C}$ is the Temporal Constraint Store, namely a set of constraint atoms in some given underlying constraint language. These constrain the time variables of the goals in $\mathcal{F}$. For example, they may specify a time window over which the time of an action can be instantiated, at execution time.

- $\Sigma$ is a set of equalities instantiating time variables with time constants. For example, when the time variables of actions are instantiated at action execution time, records of the instantiations are kept in $\Sigma$.

## 2.2 Reasoning Capabilities

KGP supports the following reasoning capabilities:

- *Planning*, which generates plans for mental goals given as input. These plans consist of temporally constrained sub-goals and actions designed for achieving the input goals.

- *Reactivity*, which is used to provide new *reactive* top-level goals, as a reaction to perceived changes in the environment and the current plans held by the agent.

- *Goal Decision*, which is used to revise the *non-reactive* top-level goals, adapting the agent's state to changes in the environment.

- *Identification of Preconditions* and *Identification of Effects* for actions, which are used to determine appropriate sensing actions for checking whether actions may be safely executed (if their preconditions are known to hold) and whether recently executed actions have been successful (by checking that some of their known effects hold).

- *Temporal Reasoning*, which allows the agent to reason about the evolving environment, and to make predictions about properties, including non-executable goals, holding in the environment, based on the (partial) information the agent acquires over its lifetime.

- *Constraint Solving*, which allows the agent to reason about the satisfiability of the temporal constraints in $\mathcal{C}$ and $\Sigma$.

In the concrete realisation of the KGP model we provide in this paper, we have chosen to realise the above capabilities in various extensions of the logic programming paradigm. In particular, we use (conventional) logic programming for Identification of Preconditions and Effects, abductive logic programming with constraints (see Section 3.2) for Planning, Reactivity and Temporal Reasoning, and logic programming with priorities (see Section 3.3) for Goal Decision.

## 2.3 Physical Capabilities

In addition to the reasoning capabilities, a KGP agent is equipped with "physical" capabilities, linking the agent to its environment, consisting of

- A *Sensing* capability, allowing the agent to observe that properties hold or do not hold, and that other agents have executed actions.

- An *Actuating* capability, for executing (physical and communicative) actions.

## 2.4 Transitions

The state $\langle KB_0, \mathcal{F}, \mathcal{C}, \Sigma \rangle$ of an agent evolves by applying transition rules, which employ the capabilities as follows:

- *Goal Introduction (GI)*, possibly changing the top-level goals in $\mathcal{F}$, and using Goal Decision.

- *Plan Introduction (PI)*, possibly changing $\mathcal{F}$ and $\mathcal{C}$ and using Planning.

- *Reactivity (RE)*, possibly changing the reactive top-level goals in $\mathcal{F}$ and $\mathcal{C}$, and using the Reactivity capability.

- *Sensing Introduction (SI)*, possibly introducing new sensing actions in $\mathcal{F}$ for checking the preconditions of actions already in $\mathcal{F}$.

- *Passive Observation Introduction (POI)*, updating $KB_0$ by recording unsolicited information coming from the environment, and using Sensing.

- *Active Observation Introduction (AOI)*, possibly updating $\Sigma$ and $KB_0$, by recording the outcome of (actively sought) sensing actions, and using Sensing.

- *Action Execution (AE)*, executing all types of actions and as a consequence updating $KB_0$ and $\Sigma$, and using Actuating.

- *State Revision (SR)*, possibly revising $\mathcal{F}$, and using Temporal Reasoning and Constraint Solving.

## 2.5 Cycle and Selection Operators

The behaviour of an agent is given by the application of transitions in sequences, repeatedly changing the state of the agent. These sequences are not determined by fixed cycles of behaviour, as in conventional agent architectures, but rather by reasoning with *cycle theories*. Cycle theories define preference policies over the order of application of transitions, which may depend on the environment and the internal state of an agent. They rely upon the use of *selection operators* for detecting which transitions are enabled and what their inputs should be, as follows:

- *action selection* for inputs to AE; this selection operator uses the Temporal Reasoning and Constraint Solving capabilities;

- *goal selection* for inputs to PI; this selection operator uses the Temporal Reasoning and Constraint Solving capabilities;

- *effect selection* for inputs to AOI; this selection operator uses the Identification of Effect reasoning capability;

- *precondition selection* for inputs to SI; this selection operator uses the Identification of Preconditions, Temporal Reasoning and Constraint Solving capabilities;

The provision of a declarative control for agents in the form of cycle theories is a highly novel feature of the model, which could, in principle, be imported into other agent systems. In the concrete realisation of the KGP model we provide in this paper, we have chosen to realise cycle theories in the same framework of logic programming with priorities and constraints (see Section 3.3) that we also use for Goal Decision.

Some of the relationships between the capabilities, transitions and the selection operators are summarised in Tables 2.5 and 2 below. Table 2.5 indicates which capabilities (rows) are used by which transitions and selection operators. Table 2 indicates which selection operators are used to compute possible inputs for which transitions in the cycle theory.

|  | *Transitions* | | | | | | | | *Selection operators* | | | |
|---|---|---|---|---|---|---|---|---|---|---|---|---|
|  | *AE* | *AOI* | *GI* | *POI* | *PI* | *RE* | *SR* | *SI* | $f_{GS}$ | $f_{AS}$ | $f_{ES}$ | $f_{PS}$ |
| `sensing` | x | x |  | x |  |  |  |  |  |  |  |  |
| `actuating` | x |  |  |  |  |  |  |  |  |  |  |  |
| $\models_{plan}$ |  |  |  |  | x |  |  |  |  |  |  |  |
| $\models_{pre}$ |  |  |  |  |  |  |  |  |  | x |  | x |
| $\models_{GD}$ |  |  | x |  |  |  |  |  |  |  |  |  |
| $\models_{react}$ |  |  |  |  |  | x |  |  |  |  |  |  |
| $\models_{TR}$ |  |  |  |  |  |  | x |  | x | x |  | x |
| $\models_{cs}$ |  |  | x |  |  | x | x |  | x | x |  | x |
| $\models_{eff}$ |  |  |  |  |  |  |  |  |  |  | x |  |

Table 1: A tabular overview of use of capabilities by transitions and selection operators. Here, $\models_{plan}$, $\models_{pre}$, $\models_{GD}$, $\models_{react}$, $\models_{TR}$, $\models_{cs}$ and $\models_{eff}$, stand for, respectively, the planning, identification of preconditions, goal decision, reactivity, temporal reasoning, constraint solving and identification of effects (reasoning) capabilities, and $f_{GS}$, $f_{AS}$, $f_{ES}$, $f_{PS}$ stand for, respectively, the goal, action, effect and precondition selection operators.

|  | *AE* | *AOI* | *GI* | *POI* | *PI* | *RE* | *SR* | *SI* |
|---|---|---|---|---|---|---|---|---|
| $f_{GS}$ |  |  |  |  | x |  |  |  |
| $f_{AS}$ | x |  |  |  |  |  |  |  |
| $f_{ES}$ |  | x |  |  |  |  |  |  |
| $f_{PS}$ |  |  |  |  |  |  |  | x |

Table 2: A tabular overview of the connections between selection operators and transitions, as required by the cycle theory. Here, $f_{GS}$, $f_{AS}$, $f_{ES}$, $f_{PS}$ stand for, respectively, the goal, action, effect and precondition selection operators.

Before we provide these components, though, we introduce below informally a scenario and some examples that will be used to illustrate the technical details of the KGP agent model throughout the paper. A full, formal presentation of these as well as additional examples will be given throughout the paper and in Section 10.

## 2.6 Examples

We draw all our examples from a ubiquitous computing scenario that we call the *San Vincenzo* scenario, presented by de Bruijn and Stathis (2003) and summarised as follows. A businessman travels for work purposes to Italy and, in order to make his trip easier, carries a personal communicator, namely a device that is a hybrid between a mobile phone and a PDA. This device is the businessman's KGP agent. This agent can be considered as a personal service agent (Mamdani, Pitt, & Stathis, 1999) (or *psa* for short) because it provides proactive information management and flexible connectivity to smart services available in the global environment within which the businessman travels within.

### 2.6.1 Setting 1

The businessman's *psa* requests from a 'San Vincenzo Station' agent, *svs*, the arrival time of the train *tr*01 from Rome. As *svs* does not have this information it answers with a refusal. Then later, *svs* receives information of the arrival time of the *tr*01 train from a 'Central Office' agent, *co*. When the *psa* requests the arrival time of *tr*01 again, *svs* will accept the request and provide the information.

This first example requires one to use the Reactivity capability to model rules of interaction and the RE transition (a) to achieve interaction amongst agents, and (b) to specify dynamic adjustments of the agent's behaviour to changes, allowing different reactions to the same request, depending on the current situation of the agent. Here, the interaction is a form of negotiation of resources amongst agents, where resources are items of information. Thus, the current situation of the agents amounts to what resources/information the agents currently own.

This example also requires the combination of transitions RE, POI, and AE to achieve the expected agents' behaviours, as follows:

1. *psa* makes the initial request by applying AE

2. *svs* becomes aware of this request by performing POI (and changing its $KB_0$ accordingly)

3. *svs* decides to reply with a refusal by performing RE (and adding the corresponding action to its plan in $\mathcal{F}$)

4. *svs* utters the refusal by performing AE

5. *svs* becomes aware, by POI, of the arrival time (modifying its $KB_0$ accordingly)

6. *psa* makes the second request by applying AE again

7. *svs* decides to reply with the requested information by performing RE (and adding the corresponding action to its plan in $\mathcal{F}$) and communicates the information by performing AE.

This sequence of transitions is given by the so-called *normal* cycle theory that we will see in Section 9.

### 2.6.2 Setting 2

In preparation of the businessman's next trip, his *psa* aims at getting a plane ticket from Madrid to Denver as well as obtaining a visa to the USA. One possible way to buy plane tickets is over the internet. Buying tickets this way is usually possible, but not to all destinations (depending on whether the airlines flying to the destinations sell tickets over the internet or not) and not without an internet connection. The *psa* does not currently have the connection, nor the information that Denver is indeed a destination for which tickets can be bought online. It plans to buy the ticket over the internet nonetheless, conditionally, but checks the conditions before executing the planned action. After successfully buying the ticket, *psa* focuses on the second goal, of obtaining a visa. This can be achieved by applying to the USA embassy in Madrid, but the application requires an address in the USA. This address can be obtained by arranging for a hotel in Denver.

This example illustrates the form of "partial" planning adopted by the KGP model (where non-executable sub-goals as well as actions may be part of plans) and shows how the combination of transition PI with SI and AE allows the *psa* agent to deal with partial information, to generate conditional plans and plans with several "layers", as follows:

1. *psa* is initially equipped with the top-level goals to get a ticket to Denver and to obtain a visa (through an earlier application of GI)

2. by PI for the first goal, *psa* adds a "partial" plan to its $\mathcal{F}$, of buying a ticket online subject to sub-goals that there is an internet connection available and that online tickets can be bought to Denver; these sub-goals are sensing goals

3. by SI, sensing actions are added to $\mathcal{F}$ to evaluate the sensing sub-goals in the environment

4. these sensing actions are executed by AE (and $KB_0$ is modified accordingly)

5. depending on the sensed values of the sensing sub-goals the buying action may or may not be executed by AE; let us assume in the remainder of the example that this action is executed

6. SR is applied to eliminate all actions (since they have already been executed), sub-goals and top-level goal of getting a ticket to Denver (since they have been achieved)

7. by PI for the remaining top-level goal of obtaining a visa, *psa* adds a plan to fill in an application form (action) and acquiring a residence address in Denver (sub-goal)

8. the action cannot be executed, as *psa* knows that the businessman is not resident in the USA; further PI introduces a plan for the sub-goal of booking a hotel (action) for the subgoal of acquiring a residence address in Denver

9. AE executes the booking action

10. AE executes the action of applying for a visa

11. SR eliminates all actions (since they have already been executed), sub-goal and top-level goal of getting a visa (since they have been achieved).

## 3. Background

In this section we give the necessary background for the reasoning capabilities and the cycle theory of KGP agents, namely:

- *Constraint Logic Programming*, pervasive to the whole model,

- *Abductive Logic Programming*, at the heart of the Planning, Reactivity and Temporal Reasoning capabilities, and

- *Logic Programming with Priorities*, at the heart of the Goal Decision capability and Cycle Theories.

### 3.1 Constraint Logic Programming

Constraint Logic Programming (CLP) (Jaffar & Maher, 1994) extends logic programming with *constraint predicates* which are not processed as ordinary logic programming predicates, defined by rules, but are checked for satisfiability and simplified by means of a built-in, "black-box" constraint solver. These predicates are typically used to constrain the values that variables in the conclusion of a rule can take (together with unification which is also treated via an equality constraint predicate). In the KGP model, constraints are used to determine the value of time variables, in goals and actions, under a suitable temporal constraint theory.

The CLP framework is defined over a structure $\Re$ consisting of a domain $D(\Re)$ and a set of constraint predicates which includes equality, together with an assignment of relations on $D(\Re)$ for each such constraint predicate. In CLP, constraints are built as first-order formulae in the usual way from primitive constraints of the form $c(t_1, \ldots, t_n)$ where $c$ is a constraint predicate symbol and $t_1, \ldots, t_n$ are terms constructed over the domain, $D(\Re)$, of values. Then the rules of a constraint logic program, $P$, take the same form as rules in conventional logic programming given by

$$H \leftarrow L_1, \ldots, L_n$$

with $H$ an (ordinary) atom, $L_1, \ldots, L_n$ literals, and $n \geq 0$. Literals can be positive, namely ordinary atoms, or negative, namely of the form *not B*, where $B$ is an ordinary atom, or constraint atoms over $\Re$. The negation symbol *not* indicates *negation as failure* (first introduced by Clark, 1978). All variables in $H$ and $L_i$ are implicitly universally quantified, with scope the entire rule. $H$ is called the *head* (or the *conclusion*) and $L_1, \ldots, L_n$ is called the *body* (or the *conditions*) of a rule of the form above. If $n = 0$, the rule is called a *fact*.

A *valuation*, $\vartheta$, of a set of variables is a mapping from these variables to the domain $D(\Re)$ and the natural extension which maps terms to $D(\Re)$. A valuation $\vartheta$, on the set of all variables appearing in a set of constraints $C$, is called an $\Re$-solution of $C$ iff $C\vartheta$, obtained by applying $\vartheta$ to $C$, is satisfied, i.e. $C\vartheta$ evaluates to true under the given interpretation of the constraint predicates and terms. This is denoted by $\vartheta \models_{\Re} C$. A set $C$ is called $\Re$-*solvable* or $\Re$-*satisfiable*, denoted by $\models_{\Re} C$, iff it has at least one $\Re$-solution, i.e. $\vartheta \models_{\Re} C$ for some valuation $\vartheta$.

One way to give the meaning of a constraint logic program $P$ is to consider the grounding of the program over its Herbrand base and all possible valuations, over $D(\Re)$, of its constraint variables. In each such rule, if the ground constraints $C$ in the body are evaluated to true then the rule is kept with the constraints $C$ dropped, otherwise the whole rule is dropped. Let $ground(P)$ be the resulting ground program. The meaning of $P$ is then given by the meaning $\models_{LP}$ of $ground(P)$, for which there are many different possible choices (Kakas, Kowalski, & Toni, 1998). The resulting overall semantics for the constraint logic program $P$ will be referred to as $\models_{LP(\Re)}$. More precisely, given a constraint logic program $P$ and a conjunction $N \wedge C$ (where $N$ is a conjunction of non-constraint literals and $C$ is a conjunction of constraint atoms), in the remainder of the paper we will write

$$P \models_{LP(\Re)} N \wedge C$$

to denote that there exists a ground substitution $\vartheta$ over the variables of $N \wedge C$ such that:

- $\vartheta \models_{\Re} C$

- $ground(P) \models_{LP} N\vartheta$.

## 3.2 Abductive Logic Programming with Constraints

An *abductive logic program* with constraints is a tuple $\langle \Re, P, A, I \rangle$ where:

- $\Re$ is a structure as in Section 3.1

- $P$ is a *constraint logic program*, namely a set of rules of the form

$$H \leftarrow L_1, \ldots, L_n$$

  as in Section 3.1

- $A$ is a set of *abducible predicates* in the language of $P$. These are predicates not occurring in the head of any clause of $P$ (without loss of generality, see (Kakas et al., 1998)). Atoms whose predicate is abducible are referred to as *abducible atoms* or simply as *abducibles*.

- $I$ is a set of *integrity constraints*, that is, a set of sentences in the language of $P$. All the integrity constraints in the KGP model have the implicative form

$$L_1, \ldots, L_n \Rightarrow A_1 \vee \ldots \vee A_m \ \ (n \geq 0, m > 0)$$

  where $L_i$ are literals (as in the case of rules) [2], $A_j$ are atoms (possibly the special atom $false$). The disjunction $A_1 \vee \ldots \vee A_m$ is referred to as the *head* of the constraint and the conjunction $L_1, \ldots, L_n$ is referred to as the *body*. All variables in an integrity constraint are implicitly universally quantified from the outside, except for variables occurring only in the *head*, which are implicitly existentially quantified with scope the head itself.

---

2. If $n = 0$, then $L_1, \ldots, L_n$ represents the special atom $true$.

Given an abductive logic program with constraints $\langle \Re, P, A, I \rangle$ and a formula (*query*) $Q$, which is an (implicitly existentially quantified) conjunction of literals in the language of $P$, the purpose of abduction is to find a (possibly minimal) set of (ground) abducible atoms $\Gamma$ which, together with $P$, "entails" (an appropriate ground instantiation of) $Q$, with respect to some notion of "entailment" that the language of $P$ is equipped with, and such that the extension of $P$ by $\Gamma$ "satisfies" $I$ (see (Kakas et al., 1998) for possible notions of integrity constraint "satisfaction"). Here, the notion of "entailment" is the combined semantics $\models_{LP(\Re)}$, as discussed in Section 3.1.

Formally, given a query $Q$, a set $\Delta$ of (possibly non-ground) abducible atoms, and a set $C$ of (possibly non-ground) constraints, the pair $(\Delta, C)$ is an *abductive answer (with constraints)* for $Q$, with respect to an abductive logic program with constraints $\langle \Re, P, A, I \rangle$, iff for all groundings $\sigma$ for the variables in $Q, \Delta, C$ such that $\sigma \models_{\Re} C$, it holds that

(i) $P \cup \Delta\sigma \models_{LP(\Re)} Q\sigma$, and

(ii) $P \cup \Delta\sigma \models_{LP(\Re)} I$, i.e. for each $B \Rightarrow H \in I$, if $P \cup \Delta\sigma \models_{LP(\Re)} B$ then $P \cup \Delta\sigma \models_{LP(\Re)} H$.

Here, $\Delta\sigma$ plays the role of $\Gamma$ in the earlier informal description of abductive answer. Note also that, by (ii), integrity constraints are not classical implications.

Note also that, when representing knowledge as an abductive logic program, one needs to decide what should go into the logic program, what in the integrity constraints and what in the abducibles. Intuitively, integrity constraints are "normative" in that they need to be enforced, by making sure that their head holds whenever their body does (by condition (ii) above), whereas logic programming rules enable, with the help of abducibles, the derivation of given goals (by condition (i) above). Finally, abducibles are chosen amongst the literals that cannot be derived by means of logic programming rules. In this paper, we will represent reactive constraints (that are condition-action rules forcing the reactive behaviour of agents) as integrity constraints, thus to some extent addressing this knowledge representation challenge posed by abductive logic programming by imposing a sort of "structure" on the abductive logic programs we use.

The notion of abductive answer can be extended to take into account an initial set of (possibly non-ground) abducible atoms $\Delta_0$ and an initial set of (possibly non-ground) constraint atoms $C_0$. In this extension, an abductive answer for $Q$, with respect to

$$(\langle \Re, P, A, I \rangle, \Delta_0, C_0)$$

is a pair $(\Delta, C)$ such that

(i) $\Delta \cap \Delta_0 = \{\}$

(ii) $C \cap C_0 = \{\}$, and

(iii) $(\Delta \cup \Delta_0, C \cup C_0)$ is an abductive answer for $Q$ with respect to $\langle \Re, P, A, I \rangle$ (in the earlier sense).

It is worth noticing that an abductive answer $(\Delta, C)$ for the query *true* with respect to

$$(\langle \Re, P, A, I \rangle, \Delta_0, C_0)$$

should be read as the fact that the abducibles in $\Delta_0 \cup \Delta$, along with the constraints in $C_0 \cup C$, guarantee the overall consistency with respect to the integrity constraints given in $I$. This will be used for the specification of some capabilities of KGP agents.

In the remainder of the paper, for simplicity, we will omit $\Re$ from abductive logic programs, which will be written simply as triples $\langle P, A, I \rangle$. In addition, all abductive logic programs that will present in KGP are variants of a core event calculus (Kowalski & Sergot, 1986), that we will define in Section 5.1.1.

## 3.3 Logic Programming with Priorities

For the purposes of this paper, a *logic program with priorities* over a constraint structure $\Re$, referred to as $\mathcal{T}$, consists of four parts:

(i) a low-level or basic part $P$, consisting of a logic program with constraints; each rule in $P$ is assigned a name, which is a term; e.g. one such rule could be

$$n(X, Y) : p(X) \leftarrow q(X, Y), r(Y)$$

with name $n(X, Y)$ naming each ground instance of the rule;

(ii) a high-level part $H$, specifying conditional, dynamic priorities amongst rules in $P$ or $H$; e.g. one such priority could be

$$h(X) : m(X) \succ n(X) \leftarrow c(X)$$

to be read: if (some instance of) the condition $c(X)$ holds, then (the corresponding instance of) the rule named by $m(X)$ should be given higher priority than (the corresponding instance of) the rule named by $n(X)$. The rule itself is named $h(X)$;

(iii) an auxiliary part $A$, which is a constraint logic program defining (auxiliary) predicates occurring in the conditions of rules in $P, H$ and not in the conclusions of any rule in $P$ or $H$;

(iv) a notion of incompatibility which, for our purposes, can be assumed to be given as a set of rules defining the predicate *incompatible*/2, e.g.

$$incompatible(p(X), p'(X))$$

to be read: any instance of the literal $p(X)$ is incompatible with the corresponding instance of the literal $p'(X)$. We assume that incompatibility is symmetric and always includes that $r \succ s$ is incompatible with $s \succ r$ for any two rule names $r, s$. We refer to the set of all incompatibility rules as $I$.

Any concrete LPP framework is equipped with a notion of entailment, which we denote by $\models_{pr}$, that is defined on top of the underlying logic programming with constraints semantics $\models_{LP(\Re)}$. This is defined differently by different approaches to LPP but they all share the following pattern. Given a logic program with priorities $\mathcal{T} = \langle P, H, A, I \rangle$ and a conjunction $\alpha$ of ground (non-auxiliary) atoms, $\mathcal{T} \models_{pr} \alpha$ iff

(i) there exists a subset $P'$ of the basic part $P$ such that $P' \cup A \models_{LP(\Re)} \alpha$, and

(ii) $P'$ is "preferred" wrt $H \cup A$ over any other subset $P''$ of $P$ that derives (under $\models_{LP(\Re)}$) a conclusion that is incompatible, wrt $I$, with $\alpha$.

Each framework has its own way of specifying what is meant for one sub-theory $P'$ to be "preferred" over another sub-theory $P''$. For example, in existing literature (Kakas et al., 1994; Prakken & Sartor, 1996; Kowalski & Toni, 1996; Kakas & Moraitis, 2003), $\models_{pr}$ is defined via argumentation. This is also the approach that we adopt, relying on the notion of an admissible argument as a sub-theory that is (i) consistent (does not have incompatible conclusions) and (ii) whose rules do not have lower priority, with respect to the high-level part $H$ of our theory, than those of any other sub-theory that has incompatible conclusions with it. The precise definition of how sets of rules are to be compared again is a matter of choice in each specific framework of LPP.

Given such a concrete definition of admissible sub-theories, the preference entailment, $\mathcal{T} \models_{pr} \alpha$, is then given by:

(i) there exists a (maximal) admissible sub-theory $\mathcal{T}'$ of $\mathcal{T}$ such that $\mathcal{T}' \models_{LP(\Re)} \alpha$, and

(ii) for any $\overline{\alpha}$ that is incompatible with $\alpha$ there does not exist an admissible sub-theory $\mathcal{T}''$ of $\mathcal{T}$ such that $\mathcal{T}'' \models_{LP(\Re)} \overline{\alpha}$.

When only the first condition of the above is satisfied we say that the theory $\mathcal{T}$ *credulously prefers or possibly prefers* $\alpha$. When both conditions are satisfied we say that the theory *sceptically prefers* $\alpha$.

## 4. The State of KGP Agents

In this Section we define formally the concept of state for a KGP agent. We also introduce all the notation that we will use in the rest of the paper in order to refer to state components. Where necessary, we will also try to exemplify our discussion with simple examples.

### 4.1 Preliminaries

In the *KGP* model we assume (possibly infinite) vocabularies of:

- *fluents*, indicated with $f, f', \ldots$,

- *action operators*, indicated with $a, a', \ldots$,

- *time variables*, indicated with $\tau, \tau', \ldots$,

- *time constants*, indicated with $t, t', \ldots, 1, 2, \ldots$, standing for natural numbers (we also often use the constant *now* to indicate the current time)

- *names of agents*, indicated with $c, c', \ldots$.

- *constants*, other than the ones mentioned above, normally indicated with lower case letters, e.g. $r, r_1, \ldots$

- a given *constraint language*, including constraint predicates $<, \leq, >, \leq, =, \neq$, with respect to some structure $\Re$ (e.g. the natural numbers) and equipped with a notion of constraint satisfaction $\models_\Re$ (see Section 3.1).

We assume that the set of fluents is partitioned into two disjoint sets:

- *mental fluents*, intuitively representing properties that the agent itself is able to plan for so that they can be satisfied, but can also be observed, and

- *sensing fluents*, intuitively representing properties which are not under the control of the agent and can only be observed by sensing the external environment.

For example, *problem_fixed* and *have_resource* may represent mental fluents, namely the properties that a (given) problem has been fixed and a (given) resource should be obtained, whereas *request_accepted* and *connection_on* may represent sensing fluents, namely the properties that a request for some (given) resource is accepted and that some (given) connection is active. Note that it is important to distinguish between mental and sensing fluents as they are treated differently by the control of the agent: mental fluents need to be planned for, whereas sensing fluents can only be observed. This will be clarified later in the paper.

We also assume that the set of action operators is partitioned into three disjoint sets:

- *physical action operators*, representing actions that the agent performs in order to achieve some specific effect, which typically causes some changes in the environment;

- *communication action operators*, representing actions which involve communications with other agents;

- *sensing action operators*, representing actions that the agent performs to establish whether some fluent (either a *sensing* fluent or an expected effect of some action) holds in the environment, or whether some agent has performed some action.

For example, $sense(connection\_on, \tau)$ is an action literal representing the act of sensing whether or not a network connection is on at time $\tau$, $do(clear\_table, \tau)$ is an action literal representing the physical action of removing every item on a given table, and $tell(c_1, c_2, request(r_1), d, \tau)$ is an action literal representing a communication action which expresses that agent $c_1$ is requesting from agent $c_2$ the resource $r_1$ within a dialogue with identifier $d$, at time $\tau$[3].

Each fluent and action operator has an associated arity: we assume that this arity is greater than or equal to 1, in that one argument (the last one, by convention) is always the time point at which a given fluent holds or a given action takes place. This time point may be a time variable or a time constant. Given a fluent $f$ of arity $n > 0$, we refer to $f(s_1, \ldots, s_{n-1}, x)$ and $\neg f(s_1, \ldots, s_{n-1}, x)$, where each $s_i$ is a constant and $x$ is a *time variable* or a *time constant* as *(timed) fluent literals*[4]. Given a fluent literal $\ell$, we denote by $\bar{\ell}$ its complement, namely $\neg f(s_1, \ldots, s_{n-1}, x)$ if $\ell$ is $f(s_1, \ldots, s_{n-1}, x)$, and $f(s_1, \ldots, s_{n-1}, x)$ if $\ell$ is $\neg f(s_1, \ldots, s_{n-1}, x)$. Examples of fluent literals are $have\_resource(pen, \tau)$, representing that a certain resource $pen$ should be obtained at some time $\tau$, as well as (the ground) $\neg on(box, table, 10)$, representing that at time 10 (a certain) box should not be on (a certain) table.

---

3. The role of the dialogue identifier will become clearer in Section 10. Intuitively, this is used to "link" communication actions occurring within the same dialogue.
4. Note that $\neg$ represents classical negation. Negation as failure occurs in the model only within the knowledge bases of agents, supporting the reasoning capabilities and the cycle theory. All other negations in the state are to be understood as classical negations.

Note that we assume that fluent literals are ground except for the time parameter. This will allow us to keep the notation simpler and to highlight the crucial role played by the time parameter. Given this simplification, we will often denote timed fluent literals simply by $\ell[x]$.

Given an action operator $a$ of arity $n > 0$, we refer to $a(s_1, \ldots, s_{n-1}, x)$, where each $s_i$ is a constant and $x$ is a *time variable* or *time constant*, as a *(timed) action literal*. Similarly to the case of fluent literals, for simplicity, we will assume that timed action literals are ground except possibly for the time. Hence, we will often denote timed action literals by $a[x]$.

We will adopt a special syntax for sensing actions, that will always have the form ($x$ is either a time variable or a time constant):

- $sense(f, x)$, where $f$ is a fluent, or

- $sense(c : a, x)$, where $c$ is the name of an agent and $a$ is an action operator.

In the first case, the sensing action allows the agent to inspect the external environment in order to check whether or not the fluent $f$ holds at the time $x$ of sensing. In the second case, the sensing action allows the agent to determine whether, at time $x$, another agent $c$ has performed some action $a$.

We will now define formally the concept of state $\langle KB_0, \mathcal{F}, \mathcal{C}, \Sigma \rangle$ of an agent.

## 4.2 Forest: $\mathcal{F}$

Each node in each tree in $\mathcal{F}$ is:

- either a *non-executable goal*, namely a (non-ground) timed fluent literal,

- or an *executable goal*, namely a (non-ground) timed action literal.

An example of a tree in $\mathcal{F}$ is given in Figure 2, where $p2$ is some given problem that the agent $(c_1)$ needs to fix by getting two resources $r_1$ and $r_2$, and where the agent has already decided to get $r_1$ from some other agent $c_2$ and has already planned to ask $c_2$ by the communication action $tell(c_1, c_2, request(r_1), d, \tau_4)$. For example, in the San Vincenzo scenario, $p2$ may be "transfer to airport needs to be arranged", $r_1$ may be a taxi, and $c2$ a taxi company, needed for transportation to some train station, and finally $r_2$ may be a train ticket.

Note that the time variable $\tau$ in non-executable goals $\ell[\tau]$ and actions $a[\tau]$ in (any tree in) $\mathcal{F}$ is to be understood as a variable that is *existentially quantified* within the whole state of the agent. Whenever a goal or action is introduced within a state, its time variable is to be understood as a distinguished, fresh variable, also serving as its *identifier*.

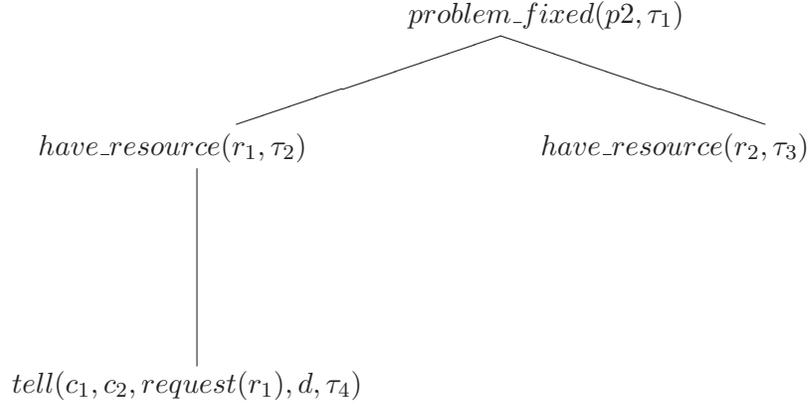

$problem\_fixed(p2, \tau_1)$

$have\_resource(r_1, \tau_2)$              $have\_resource(r_2, \tau_3)$

$tell(c_1, c_2, request(r_1), d, \tau_4)$

Figure 2: An example tree in $\mathcal{F}$

As indicated in Section 2, roots of trees are referred to as *top-level goals*, executable goals are often called simply *actions*, non-executable goals may be top-level goals or *sub-goals*. For example, in Figure 2, the node with identifier $\tau_1$ is a top-level goal, the nodes with identifiers $\tau_2, \tau_3$ are sub-goals and the node with identifier $\tau_4$ is an action.

**Notation 4.1** Given a forest $\mathcal{F}$ and a tree $\mathcal{T} \in \mathcal{F}$:

- for any node $n$ of $\mathcal{T}$, $parent(n, \mathcal{T})$, $children(n, \mathcal{T})$, $ancestors(n, \mathcal{T})$, $siblings(n, \mathcal{T})$, $descendents(n, \mathcal{T})$, will indicate the parent of node $n$ in $\mathcal{T}$, the children of $n$ in $\mathcal{T}$, etc. and $leaf(n, \mathcal{T})$ will have value true if $n$ is a leaf in $\mathcal{T}$, false otherwise.

- for any node $n$ of $\mathcal{F}$, $parent(n, \mathcal{F})$, $children(n, \mathcal{F})$, $ancestors(n, \mathcal{F})$, $siblings(n, \mathcal{F})$, $descendents(n, \mathcal{F})$, $leaf(n, \mathcal{F})$ will indicate the $parent(n, \mathcal{T})$ for the tree $\mathcal{T}$ in $\mathcal{F}$ where $n$ occurs, etc. ($\mathcal{T}$ is unique, due to the uniqueness of the time variable identifying nodes).

- $nodes(\mathcal{T})$ will represent the set of nodes in $\mathcal{T}$, and $nodes(\mathcal{F})$ will represent the set $nodes(\mathcal{F}) = \bigcup_{\mathcal{T} \in \mathcal{F}} nodes(\mathcal{T})$.

Again, as indicated in Section 2, each top-level goal in each tree in $\mathcal{F}$ will be either *reactive* or *non-reactive*. We will see, in Section 7, that reactive top-level goals are introduced into the state by the RE transition whereas non-reactive top-level goals are introduced by the GI transition. For example, $\mathcal{F}$ of agent $c_1$ may consist of the tree in Figure 2, with root a non-reactive goal, as well as a tree with root the reactive goal (action) $tell(c_1, c_2, accept\_request(r_3), d', \tau_5)$. This action may be the reply (planned by agent $c_1$) to some request for resource $r_3$ by agent $c_2$ (for example, in the San Vincenzo scenario, $r_3$ may be a meeting requested by some colleague).

**Notation 4.2** Given a forest $\mathcal{F}$

- $Roots^r(\mathcal{F})$ (resp. $Roots^{nr}(\mathcal{F})$) will denote the set of all reactive (resp. non-reactive) top-level goals in $\mathcal{F}$

- $nodes^r(\mathcal{F})$ (resp. $nodes^{nr}(\mathcal{F})$) will denote the subset of $nodes(\mathcal{F})$ consisting of nodes in all trees whose root is in $Roots^r(\mathcal{F})$ (resp. $Roots^{nr}(\mathcal{F})$)

- $r(\mathcal{F})$ (resp. $nr(\mathcal{F})$) stands for the *reactive (resp. non-reactive) part of* $\mathcal{F}$, namely the set of all trees in $\mathcal{F}$ whose root is in $Roots^r(\mathcal{F})$ (resp. $Roots^{nr}(\mathcal{F})$).

Trivially, $r(\mathcal{F})$ and $nr(\mathcal{F})$ are disjoint, and $\mathcal{F} = r(\mathcal{F}) \cup nr(\mathcal{F})$.

### 4.3 Temporal Constraint Store: $\mathcal{C}$

This is a set of constraint atoms, referred to as *temporal constraints*, in the given underlying constraint language. Temporal constraints refer to time constants as well as to time variables associated with goals (currently or previously) in the state.

For example, given a forest with the tree in Figure 2, $\mathcal{C}$ may contain $\tau_1 > 10, \tau_1 \leq 20$, indicating that the top-level goal (of fixing problem $p2$) needs to be achieved within the time interval $(10, 20]$, $\tau_2 < \tau_1, \tau_3 < \tau_1$, indicating that resources $r_1$ and $r_2$ need to be acquired before the top-level goal can be deemed to be achieved, and $\tau_4 < \tau_2$, indicating that the agent needs to ask agent $c_2$ first. Note that we do not need to impose that $\tau_2$ and $\tau_3$ are executed in some order, namely $\mathcal{C}$ may contain neither $\tau_2 < \tau_3$, nor $\tau_3 < \tau_2$.

### 4.4 Agents' Dynamic Knowledge Base: $KB_0$

$KB_0$ is a set of logic programming facts in the state of an agent, recording the actions which have been executed (by the agent or by others) and their time of execution, as well as the properties (i.e. fluents and their negation) which have been observed and the time of the observation. Formally, these facts are of the following forms:

- $executed(a, t)$ where $a[t]$ is a ground action literal, meaning that action $a$ has been executed by the agent at time $t$.

- $observed(\ell, t)$ where $\ell[t]$ is a ground fluent literal, meaning that $\ell$ has been observed to hold at time $t$.

- $observed(c, a[t'], t)$ where $c$ is an agent's name, different from the name of the agent whose state we are defining, $t$ and $t'$ are time constants, and $a[t']$ is a (ground) action literal. This means that the given agent has observed at time $t$ that agent $c$ has executed the action $a$ at time $t'$ [5].

---

5. We will see that, by construction, it will always be the case that $t' \leq t$. Note that the time of executed actions, $t'$, and the time of their observation, $t$, will typically be different in any concrete implementation of the KGP model, as they depend, for example, on the time of execution of transitions within the operational trace of an agent.

Note that all facts in $KB_0$ are variable-free, as no time variables occur in them. Facts of the first kind record actions that have been executed by the agent itself. Facts of the second kind record observations made by the agent in the environment, excluding actions executed by other agents, which are represented instead as facts of the third kind.

For example, if the action labelled by $\tau_4$ in Figure 2 is executed (by the AE transition) at time 7 then $executed(tell(c_1, c_2, request(r_1), d), 7)$ will be added to $KB_0$. Moreover, if, at time 9, $c_1$ observes (e.g. by transition POI) that it has resource $r_2$, then the observation $observed(have\_resource(r_2), 9)$ will be added to $KB_0$. Finally, $KB_0$ may contain

$$observed(c_2, tell(c_2, c_1, request(r_3), d', 1), 6)$$

to represent that agent $c_1$ has become aware, at time 6, that agent $c_2$ has requested, at the earlier time 1, resource $r_3$ from $c_1$.

### 4.5 Instantiation of Time Variables: $\Sigma$

When a time variable $\tau$ occurring in some non-executable goal $\ell[\tau]$ or some action $a[\tau]$ in $\mathcal{F}$ is instantiated to a time constant $t$ (e.g. at action execution time), the actual instantiation $\tau = t$ is recorded in the $\Sigma$ component of the state of the agent. For example, if the action labelled by $\tau_4$ in Figure 2 is executed at time 7, then $\tau_4 = 7$ will be added to $\Sigma$.

The use of $\Sigma$ allows one to record the instantiation of time variables while at the same time keeping different goals with the same fluent distinguished. Clearly, for each time variable $\tau$ there exists at most one equality $\tau = t$ in $\Sigma$.

**Notation 4.3** Given a time variable $\tau$, we denote by $\Sigma(\tau)$ the time constant $t$, if any, such that $\tau = t \in \Sigma$.

It is worth pointing out that the valuation of any temporal constraint $c \in \mathcal{C}$ will always take the equalities in $\Sigma$ into account. Namely, any ground valuation for the temporal variables in $c$ must agree with $\Sigma$ on the temporal variables assigned to them in $\Sigma$. For example, given $\Sigma = \{\tau = 3\}$ and $\mathcal{C} = \{\tau_1 > \tau\}$, then $\tau_1 = 10$ is a suitable valuation, whereas $\tau_1 = 1$ is not.

## 5. Reasoning Capabilities

In this section, we give detailed specifications for the various reasoning capabilities, specified within the framework of ordinary logic programming (for Temporal Reasoning and Identification of Preconditions and Effects), of Abductive Logic Programming with Constraints (Section 3.2, for Planning and Reactivity), of Logic Programming with Priorities with Constraints (Section 3.3, for Goal Decision), of constraint programming (Section 3.1, for Constraint Solving).

The reasoning capabilities are defined by means of a notion of "entailment" with respect to an appropriate knowledge base (and a time point $now$, where appropriate), as follows:

- $\models_{TR}$ and $KB_{TR}$ for Temporal Reasoning, where $KB_{TR}$ is a constraint logic program and a variant of the framework of the Event Calculus (EC) for reasoning about actions, events and changes (Kowalski & Sergot, 1986) [6];

- $\models_{plan}^{now}$ and $KB_{plan}$ for Planning, where $KB_{plan}$ is an abductive logic program with constraints, extending $KB_{TR}$;

- $\models_{react}^{now}$ and $KB_{react}$ for Reactivity, where $KB_{react}$ is an extension of $KB_{plan}$, incorporating additional integrity constraints representing reactive rules;

- $\models_{pre}$ and $KB_{pre}$, where $KB_{pre}$ is a logic program contained in $KB_{TR}$;

- $\models_{eff}$ and $KB_{eff}$, where $KB_{eff}$ is a logic program contained in $KB_{TR}$;

- $\models_{GD}^{now}$ and $KB_{GD}$, where $KB_{GD}$ is a logic program with priorities and constraints.

The constraint solving capability is defined in terms of an "entailment" $\models_{cs}$ which is basically $\models_{\Re}$ as defined in Section 3.1.

## 5.1 Temporal Reasoning, Planning, Reactivity, Identification of Preconditions and Effects: EC-based Capabilities

These reasoning capabilities are all specified within the framework of the event calculus (EC) for reasoning about actions, events and changes (Kowalski & Sergot, 1986). Below, we first give the core EC and then show how to use it to define the various capabilities in this section.

### 5.1.1 Preliminaries: Core Event Calculus

In a nutshell, the EC allows one to write meta-logic programs which "talk" about object-level concepts of *fluents*, *events* (that we interpret as action *operators*) [7], and *time points*. The main meta-predicates of the formalism are:

- $holds\_at(F, T)$ - a fluent $F$ holds at a time $T$;

- $clipped(T_1, F, T_2)$ - a fluent $F$ is clipped (from holding to not holding) between times $T_1$ and $T_2$;

- $declipped(T_1, F, T_2)$ - a fluent $F$ is declipped (from not holding to holding) between times $T_1$ and $T_2$;

- $initially(F)$ - a fluent $F$ holds at the initial time, say time 0;

- $happens(O, T)$ - an operation $O$ happens at a time $T$;

- $initiates(O, T, F)$ - a fluent $F$ starts to hold after an operation $O$ at time $T$;

---

6. A more sophisticated, abductive logic programming version of $\models_{TR}$ and $KB_{TR}$ is given by Bracciali and Kakas (2004).

7. In this section we use the original event calculus terminology of events instead of operators, as in the rest of the paper.

- $terminates(O, T, F)$ - a fluent $F$ ceases to hold after an operation $O$ at time $T$.

Roughly speaking, the last two predicates represent the cause-effects links between operations and fluents in the modelled world. We will also use a meta-predicate

- $precondition(O, F)$ - the fluent $F$ is one of the preconditions for the executability of the operation $O$.

Fluent literals in an agent's state are mapped onto the EC as follows. The EC-like representation of a fluent literal $f[\tau]$ (resp. $\neg f[\tau]$) in an agent's state is the atom $holds\_at(f, \tau)$ (resp. $holds\_at(\neg f, \tau)$). Moreover, when arguments other than the time variable need to be considered, the EC representation of a fluent literal $f(x_1, \ldots, x_n, \tau)$ (resp. $\neg f(x_1, \ldots, x_n, \tau)$) is $holds\_at(f(x_1, \ldots, x_n), \tau)$ (resp. $holds\_at(\neg f(x_1, \ldots, x_n), \tau)$. [8]

Similarly, action literals in the state of an agent can be represented in the EC in a straightforward way. Given an action literal $a[\tau]$ its EC representation is $happens(a, \tau)$. When arguments other than time are considered, as e.g. in $a(x_1, \ldots, x_n, \tau)$, the EC representation is given by $happens(a(x_1, \ldots x_n), \tau)$.

In the remainder of the paper, with an abuse of terminology, we will sometimes refer to $f(x_1, \ldots, x_n)$ and $\neg f(x_1, \ldots, x_n)$ interchangeably as fluent literals or fluents (although strictly speaking they are fluent literals), and to $a(x_1, \ldots x_n)$ interchangeably as action literals or action operators (although strictly speaking they are action literals).

The EC allows one to represent a wide variety of phenomena, including operations with indirect effects, non-deterministic operations, and concurrent operations (Shanahan, 1997).

The core EC we use in this paper consists of two parts: domain-independent rules and domain-dependent rules. The basic *domain-independent rules*, directly borrowed from the original EC, are:

$$
\begin{aligned}
holds\_at(F, T_2) \leftarrow \quad & happens(O, T_1), initiates(O, T_1, F), \\
& T_1 < T_2, not\ clipped(T_1, F, T_2) \\
holds\_at(\neg F, T_2) \leftarrow \quad & happens(O, T_1), terminates(O, T_1, F), \\
& T_1 < T_2, not\ declipped(T_1, F, T_2) \\
holds\_at(F, T) \leftarrow \quad & initially(F), 0 \leq T, not\ clipped(0, F, T) \\
holds\_at(\neg F, T) \leftarrow \quad & initially(\neg F), 0 \leq T, not\ declipped(0, F, T) \\
clipped(T_1, F, T_2) \leftarrow \quad & happens(O, T), terminates(O, T, F), T_1 \leq T < T_2 \\
declipped(T_1, F, T_2) \leftarrow \quad & happens(O, T), initiates(O, T, F), T_1 \leq T < T_2
\end{aligned}
$$

The *domain-dependent rules* define $initiates$, $terminates$, and $initially$, e.g. in the case of setting 2.6.1 in Section 2.6 we may have

$$
\begin{aligned}
initiates(tell(C, svs, & inform(Q, I), D), T, have\_info(svs, Q, I)) \leftarrow \\
& holds\_at(trustworthy(C), T) \\
initially(\neg have\_info(svs, & arrival(tr01), I)
\end{aligned}
$$

---

8. Note that we write $holds\_at(\neg f(x_1, \ldots, x_n), \tau)$ instead of $not\ holds\_at(f(x_1, \ldots, x_n), \tau)$, as done e.g. by Shanahan, 1997, because we want to reason *at the object-level* about properties being true or false in the environment. We use *not* within the *meta-level* axioms of the event calculus (see below) to implement persistence.

$$initially(trustworthy(co))$$

Namely, an action by agent $C$ of providing information $I$ concerning a query $Q$ to the agent $svs$ (the "San Vincenzo station" agent) initiates the agent $svs$ having the information about $Q$, provided that $C$ is trustworthy. Moreover, initially agent $co$ (the "Central Office" agent) is trustworthy, and agent $svs$ has no information about the arrival time of $tr01$. The conditions for the rule defining *initiates* can be seen as preconditions for the effects of the operator *tell* to take place. Preconditions for the executability of operators are specified by means of a set of rules (facts) defining the predicate *precondition*, e.g.

$$precondition(tell(svs, C, inform(Q, I), D), have\_info(svs, Q, I))$$

namely the precondition for agent $svs$ to inform any agent $C$ of $I$ about $Q$ is that $svs$ indeed has information $I$ about $Q$.

Notice that the presence in the language of fluents and their negation, e.g. $f$ and $\neg f$, poses the problem of "inconsistencies", i.e. it may be the case that both $holds\_at(f, t)$ and $holds\_at(\neg f, t)$ can be derived from the above axioms and a set of events (i.e. a given set of *happens* atoms). However, it can easily be shown that this is never the case, provided that the domain-dependent part does not contain two conflicting statements of the form $initially(f)$ and $initially(\neg f)$ since inconsistencies cannot be caused except at the initial time point (see e.g. Miller & Shanahan, 2002, p. 459).

In the remainder of the paper we will assume that the domain-dependent part is always consistent for our agents.

To allow agents to draw conclusions from the contents of $KB_0$, which represents the "narrative" part of the agent's knowledge, we add to the domain-independent rules the following *bridge rules*:

$$
\begin{aligned}
holds\_at(F, T_2) &\leftarrow & observed(F, T_1), T_1 \leq T_2, not\ clipped(T_1, F, T_2) \\
holds\_at(\neg F, T_2) &\leftarrow & observed(\neg F, T_1), T_1 \leq T_2, not\ declipped(T_1, F, T_2) \\
happens(O, T) &\leftarrow & executed(O, T) \\
happens(O, T) &\leftarrow & observed(\_, O[T], \_)
\end{aligned}
$$

Notice that these bridge rules make explicit the translation from the state representation to the EC representation of fluents and actions we have mentioned earlier on in this section. Note also that we assume that a fluent holds from the time it is observed to hold. This choice is dictated by the rationale that observations can only be considered and reasoned upon from the moment the agent makes them. On the other hand, actions by other agents have effect from the time they have been executed [9].

Having introduced the ability to reason with narratives of events and observations, we need to face the problem of "inconsistency" due to conflicting observations, e.g. an agent may observe that both a fluent and its negation hold at the same time. As we have done above for the set of *initially* atoms, we will assume that the external world is consistent too, i.e. it can never happen that both $observed(f, t)$ and $observed(\neg f, t)$ belong to $KB_0$, for any fluent $f$ and time point $t$.

---

9. If the time of the action is unknown at observation time, then the last rule above may be replaced by
$$happens(O, T) \leftarrow observed(\_, O[\_], T)$$
namely the value of a fluent is changed according to observations from the moment the observations are made.

However, we still need to cope with the frame consistency problem, which arises, e.g. given observations $observed(f, t)$ and $observed(\neg f, t')$, with $t \neq t'$. This issue is analogous to the case when two different events happen at the same time point and they initiate and terminate the same fluent. In the original EC suitable axioms for the predicates clipped and declipped are added, as given above, to avoid both a fluent and its negation holding at the same time after the happening of two such events at the same time. We adopt here a similar solution to cope with observations, namely by adding the following two axioms to the domain-independent part:

$$clipped(T_1, F, T_2) \leftarrow \quad observed(\neg F, T), T_1 \leq T < T_2$$
$$declipped(T_1, F, T_2) \leftarrow \quad observed(F, T), T_1 \leq T < T_2$$

This solution may be naive in some circumstances and more sophisticated solutions may be adopted, as e.g. the one proposed by Bracciali and Kakas (2004).

### 5.1.2 Temporal Reasoning

The temporal reasoning capability is invoked by other components of the KGP model (namely the Goal Decision capability, the State Revision transition and some of the selection operators, see Section 7) to prove or disprove that a given (possibly temporally constrained) fluent literal holds, with respect to a given theory $KB_{TR}$. For the purposes of this paper $KB_{TR}$ is an EC theory composed of the domain-independent and domain-dependent parts as given in Section 5.1.1, and of the "narrative" part given by $KB_0$. Then, given a state $S$, a fluent literal $\ell[\tau]$ and a possibly empty set [10] of temporal constraints $TC$, the temporal reasoning capability $\models_{TR}$ is defined as

$$S \models_{TR} \ell[\tau] \wedge TC \text{ iff } KB_{TR} \models_{LP(\Re)} holds\_at(\ell, \tau) \wedge TC.$$

For example, given the EC formulation in Section 5.1.1 for setting 2.6.1 in Section 2.6, if the state $S = \langle KB_0, \mathcal{F}, \mathcal{C}, \Sigma \rangle$ for agent $svs$ contains

$$KB_0 = \{observed(co, tell(co, svs, inform(arrival(tr01), 18), d, 15), 17)\},$$

then $S \models_{TR} have\_info(svs, arrival(tr01), 18, \tau) \wedge \tau > 20$.

### 5.1.3 Planning

A number of abductive variants of the EC have been proposed in the literature to deal with planning problems, e.g. the one proposed by Shanahan, 1989. Here, we propose a novel variant, somewhat inspired by the $\mathcal{E}$-language (Kakas & Miller, 1997), to allow situated agents to generate partial plans in a dynamic environment.

We will refer to $KB_{plan} = \langle P_{plan}, A_{plan}, I_{plan} \rangle$ as the abductive logic program where:

---

10. Here and in the remainder of the paper sets are seen as conjunctions, where appropriate.

- $A_{plan} = \{assume\_holds, assume\_happens\}$, namely we consider two abducible predicates, corresponding to assuming that a fluent holds or that an action occurs, respectively, at a certain time point;

- $P_{plan}$ is obtained by adding to the core EC axioms and the "narrative" given by $KB_0$ the following rules

  $happens(O, T) \leftarrow assume\_happens(O, T)$
  $holds\_at(F, T) \leftarrow assume\_holds(F, T)$

- $I_{plan}$ contains the following set of integrity constraints

  $holds\_at(F, T), holds\_at(\neg F, T) \Rightarrow false$
  $assume\_happens(O, T), precondition(O, P) \Rightarrow holds\_at(P, T)$
  $assume\_happens(O, T), not \; executed(O, T), time\_now(T') \Rightarrow T > T'$

These integrity constraints in $I_{plan}$ prevent the generation of (partial) plans which are unfeasible. The first integrity constraint makes sure that no plan is generated which entails that a fluent and its negation hold at the same time. The second integrity constraint makes sure that, if a plan requires an action to occur at a certain time point, the further goal of enforcing the preconditions of that action to hold at that time point is taken into account in the same plan. This means that, if those preconditions are not already known to hold, the plan will need to accommodate actions to guarantee that they will hold at the time of execution of the action. Finally, the last integrity constraint forces all assumed unexecuted actions in a plan to be executable in the future, where the predicate $time\_now(\_)$ is meant to return the current time.

It is worth recalling that, in concrete situations, $P_{plan}$ and $I_{plan}$ will also contain domain-dependent rules and constraints. Domain-dependent rules may be needed not only to define *initiates*, *terminates*, *initially* and *precondition*, but they may also contain additional rules/integrity constraints expressing *ramifications*, e.g.

$holds\_at(f, T) \Rightarrow holds\_at(f_1, T) \lor \ldots \lor holds\_at(f_n, T)$

for some specific fluents in the domain. Moreover, integrity constraints may represent specific properties of actions and fluents in the domain. As an example, a domain-dependent constraint could express that two actions of some type cannot be executed at the same time, e.g.

$holds\_at(tell(c, X, accept\_request(R), D), T),$
$holds\_at(tell(c, X, refuse\_request(R), D), T) \Rightarrow false$

Intuitively, constructing a (partial) plan for a goal (that is a given leaf node in the current forest) amounts to identifying actions and further sub-goals allowing to achieve the goal, while assuming that all other nodes in the forest, both executable and non-executable, are feasible. Concretely, the abductive logic program $KB_{plan}$ supports partial planning as follows. Whenever a plan for a given goal requires the agent to execute an action, $a[\tau]$ say, the corresponding atom $assume\_happens(a, \tau)$ is assumed, which amounts to intending to execute the action (at some concrete time instantiating $\tau$). On the other hand, if a plan for a given goal requires to plan for a sub-goal, $\ell[\tau]$ say, the corresponding atom $assume\_holds(\ell, \tau)$ may be assumed, which amounts to setting the requirement that further planning will be needed for the sub-goal itself. Notice that if only *total* plans are taken into account, no atoms of the form $assume\_holds(\_, \_)$ will ever be generated.

Formally, let $KB_{plan}^{now}$ be $KB_{plan} \cup \{time\_now(now)\}$, where $now$ is a time constant (intuitively, the time when the planning capability is invoked). Then, the planning capability $\models_{plan}^{now}$ is specified as follows [11].

Let $S = \langle KB_0, \mathcal{F}, \mathcal{C}, \Sigma \rangle$ be a state, and $G = \ell[\tau]$ be a mental goal labeling a leaf node in a tree $\mathcal{T}$ of $\mathcal{F}$. Let also

$$\mathcal{CA} = \{assume\_happens(a, \tau') \mid a[\tau'] \in nodes(\mathcal{F})\},$$

$$\mathcal{CG} = \{assume\_holds(\ell', \tau') \mid \ell'[\tau'] \in nodes(\mathcal{F}) \setminus \{\ell[\tau]\}\}$$

and

- $\Delta_0 = \mathcal{CA} \cup \mathcal{CG}$

- $C_0 = \mathcal{C} \cup \Sigma$.

Then,

$$S, G \models_{plan}^{now} (\mathcal{X}_s, TC)$$

iff

$$\mathcal{X}_s = \{a[\tau'] \mid assume\_happens(a, \tau') \in \Delta\} \cup \{\ell'[\tau'] \mid assume\_holds(\ell', \tau') \in \Delta\}$$

for some $(\Delta, TC)$ which is an abductive answer for $holds\_at(\ell, \tau)$, wrt $(KB_{plan}^{now}, \Delta_0, C_0)$. If no such abductive answer exists, then $S, G \models_{plan}^{now} \bot$, where $\bot$ is used here to indicate failure (i.e. that no such abductive answer exists).

As an example, consider setting 2.6.2 in Section 2.6. The domain-dependent part of $KB_{plan}$ for agent $psa$ (looking after the businessman in our scenario) contains

$initiates(buy\_ticket\_online(From, To), T, have\_ticket(From, To))$
$precondition(buy\_ticket\_online(From, To), available\_connection)$
$precondition(buy\_ticket\_online(From, To), available\_destination(To))$

The goal $G$ is $have\_ticket(madrid, denver, \tau)$. Assume $\mathcal{F}$ only consists of a single tree consisting solely of the root $G$, thus $\mathcal{CA} = \mathcal{CG} = \{\}$. Then, $S, G \models_{plan}^{now} (\mathcal{X}_s, TC)$ where

$\mathcal{X}_s = \{buy\_ticket\_online(madrid, denver, \tau'),$
$\quad available\_connection(\tau''), available\_destination(denver, \tau''')\}$

and $TC = \{\tau' < \tau, \tau' = \tau'' = \tau''', \tau' > now\}$.

### 5.1.4 Reactivity

This capability supports the reasoning of reacting to stimuli from the external environment as well as to decisions taken while planning.

As knowledge base $KB_{react}$ supporting reactivity we adopt an extension of the knowledge base $KB_{plan}$ as follows. $KB_{react} = \langle P_{react}, A_{react}, I_{react} \rangle$ where

- $P_{react} = P_{plan}$

---

11. For simplicity we present the case of planning for single goals only.

- $A_{react} = A_{plan}$

- $I_{react} = I_{plan} \cup \mathcal{RR}$

where $\mathcal{RR}$ is a set of *reactive constraints*, of the form

$$Body \Rightarrow Reaction, TC$$

where

- *Reaction* is either $assume\_holds(\ell, T)$, $\ell[T]$ being a timed fluent literal, or $assume\_happens(a, T)$, $a[T]$ being a timed action literal, [12] and

- *Body* is a non-empty conjunction of items of the form (where $\ell[X]$ is a timed fluent literal and $a[X]$ is a timed action literal, for any $X$):

  (i) $observed(\ell, T')$,

  (ii) $observed(c, a[T'], T'')$,

  (iii) $executed(a, T')$,

  (iv) $holds\_at(\ell, T')$,

  (v) $assume\_holds(\ell, T')$,

  (vi) $happens(a, T')$,

  (vii) $assume\_happens(a, T')$,

  (viii) temporal constraints on (some of) $T, T', T''$

  which contains at least one item from one of (i), (ii) or (iii).

- $TC$ are temporal constraints on (some of) $T, T', T''$.

As for integrity constraints in abductive logic programming, all variables in *Body* are implicitly universally quantified over the whole reactive constraint, and all variables in *Reaction*, $TC$ not occurring in *Body* are implicitly existentially quantified on the righthand side of the reactive constraint. [13]

Notice that *Body* must contain at least a *trigger*, i.e. a condition to be evaluated in $KB_0$. Intuitively, a reactive constraint $Body \Rightarrow Reaction, TC$ is to be interpreted as follows: if (some instantiation of) all the observations in *Body* hold in $KB_0$ and (some corresponding instantiation of) all the remaining conditions in *Body* hold, then (the appropriate instantiation of) *Reaction*, with associated (the appropriate instantiation of) the temporal constraints $TC$, should be added to $\mathcal{F}$ and $\mathcal{C}$, respectively. Notice that *Reaction* is an abducible so that no planning is performed by the reactivity capability.

---

12. Here and below, with an abuse of notation, we use the notions of timed fluent and action literals liberally and allow them to be non-ground, even though we have defined timed fluent and action literals as ground except possibly for the time parameter.

13. Strictly speaking, syntactically reactive constraints are not integrity constraints (due to the presence of a conjunction, represented by ",", rather than a disjunction in the head). However, any reactive constraint $Body \Rightarrow Reaction, TC$ can be transformed into an integrity constraint $Body \Rightarrow New$ with a new clause $New \leftarrow Reaction, TC$ in $P_{react}$. Thus, with an abuse of notation, we treat reactive constraints as integrity constraints.

Formally, let $KB_{react}^{now}$ be the theory $KB_{react} \cup \{time\_now(now)\}$, where $now$ is a time constant (intuitively, the time when the capability is invoked). Then, the reactivity capability $\models_{react}^{now}$ is specified as follows. Let $S = \langle KB_0, \mathcal{F}, \mathcal{C}, \Sigma \rangle$ be a state. Let

$$\mathcal{CA} = \{assume\_happens(a, \tau) \mid a[\tau] \in nodes^{nr}(\mathcal{F})\},$$

$$\mathcal{CG} = \{assume\_holds(\ell, \tau) \mid \ell[\tau] \in nodes^{nr}(\mathcal{F})\}$$

and

- $\Delta_0 = \mathcal{CA} \cup \mathcal{CG}$

- $C_0 = \mathcal{C} \cup \Sigma$.

Then,

$$S \models_{react}^{now} (\mathcal{X}_s, TC)$$

iff

$$\mathcal{X}_s = \{a[\tau] \mid assume\_happens(a, \tau) \in \Delta\} \cup \{\ell[\tau] \mid assume\_holds(\ell, \tau) \in \Delta\}$$

for some $(\Delta, TC)$ which is an abductive answer for the query $true$ wrt $(KB_{react}^{now}, \Delta_0, C_0)$. If no such abductive answer exists, then $S \models_{react}^{now} \bot$, where $\bot$ is used here to indicate failure (i.e. that no such abductive answer exists).

As an example, consider setting 2.6.1 in Section 2.6, and $KB_{plan}$ as given in Sections 5.1.1 and 5.1.3. Let $\mathcal{RR}$ of agent $svs$ consist of:

$observed(C, tell(C, svs, request(Q), D, T0), T), \ holds\_at(have\_info(svs, Q, I), T)$
$\quad \Rightarrow assume\_happens(tell(svs, C, inform(Q, I), D), T'), \ T' > T$

$observed(C, tell(C, svs, request(Q), D, T0), T), \ holds\_at(no\_info(svs, Q), T)$
$\quad \Rightarrow assume\_happens(tell(svs, C, refuse(Q), D), T'), \ T' > T$

Then, given $now = 30$ and $S = \langle KB_0, \mathcal{F}, \mathcal{C}, \Sigma \rangle$ with

$KB_0 = \{observed(co, tell(co, svs, inform(arrival(tr01), 18), d1, 15), 17),$
$\qquad observed(psa, tell(psa, svs, request(arrival(tr01)), d2, 20), 22)\}$

we obtain

$\quad S \models_{react}^{now} (\{tell(svs, psa, inform(arrival(tr01), 18), d2, \tau)\}, \tau > 30).$

### 5.1.5 Identification Of Preconditions

This capability is used by KGP agents to determine the preconditions for the executability of actions which are planned for. These preconditions are defined in the domain-dependent part of the EC by means of a set of rules of the form $precondition(O, F)$, representing that the fluent $F$ is a precondition for the executability of an action with action operator $O$ (see 5.1.1). Let $KB_{pre}$ be the subset of $KB_{TR}$ containing the rules defining $precondition(\_, \_)$.

Then the identification of preconditions capability $\models_{pre}$ is specified as follows. Given a state $S = \langle KB_0, \mathcal{F}, \mathcal{C}, \Sigma \rangle$ and a timed action literal $a[\tau]$

$$S, a[\tau] \models_{pre} Cs$$

iff

$$Cs = \bigwedge \{\ell[\tau] \mid KB_{pre} \models_{LP} precondition(a, \ell)\}^{14}.$$

### 5.1.6 IDENTIFICATION OF EFFECTS

This capability is used by KGP agents to determine the effects of actions that have already been executed, in order to check whether these actions have been successful. Note that actions may have been unsuccessful because they could not be executed, or were executed but they did not have the expected effect. Both are possible in situations where the agent does not have full knowledge about the environment in which it is situated.

These effects are defined in the domain-dependent part of the EC by means of the set of rules defining the predicates *initiates* and *terminates*. Let $KB_{eff}$ be the theory consisting of the domain-dependent and domain-independent parts of the EC, as well as the narrative part $KB_0$. Then, the identification of effects $\models_{eff}$ is specified as follows. Given a state $S = \langle KB_0, \mathcal{F}, \mathcal{C}, \Sigma \rangle$ and an action operator $a[t]$,

$$S, a[t] \models_{eff} \ell$$

iff

- $\ell = f$ and $KB_{eff} \models_{LP} initiates(a, t, f)$

- $\ell = \neg f$ and $KB_{eff} \models_{LP} terminates(a, t, f)$

## 5.2 Constraint Solving

The Constraint Solving capability can be simply defined in terms of the structure $\Re$ and the $\models_{\Re}$ notion presented in Section 3.1. Namely, given a state $S = \langle KB_0, \mathcal{F}, \mathcal{C}, \Sigma \rangle$ and a set of constraints $TC$:

- $S \models_{cs} TC$ iff $\models_{\Re} \mathcal{C} \wedge \Sigma \wedge TC$;

- there exists a total valuation $\sigma$ such that $S, \sigma \models_{cs} TC$ iff there exists a total valuation $\sigma$ such that $\sigma \models_{\Re} \mathcal{C} \wedge \Sigma \wedge TC$.

## 5.3 Goal Decision

The Goal Decision reasoning capability allows the agent to decide, at a given time point, the (non-reactive) top-level goals to be pursued, for which it will then go on to generate plans aiming at achieving them. The generated goals are the goals of current preferred interest but this interest may change over time.

---

14. We assume that $\bigwedge \{\} = true$.

The Goal Decision capability operates according to a theory, $KB_{GD}$, in which the agent represents its goal preference policy. $KB_{GD}$ includes $KB_{TR}$ and thus the dynamic, observed knowledge, $KB_0$, in the current state of the agent. $KB_{GD}$ is expressed in a variant of LPP described in Section 3.3, whereby the rules in the lower or basic part $P$ of the LPP theory $\mathcal{T}$ have the form ($\underline{T}$ being a possibly empty sequence of variables):

$$n(\tau, \underline{T}) : G[\tau, \underline{T}] \leftarrow B[\underline{T}], C[\underline{T}]$$

where

- $\tau$ is a time variable, existentially quantified with scope the head of the rule and not a member of $\underline{T}$;

- all variables except for $\tau$ are universally quantified with scope the rule;

- the head $G[\tau, \underline{T}]$ of the rule consists of a fluent literal conjoined with a (possibly empty) set of temporal constraints, represented as $\langle \ell[\tau], TC[\tau, \underline{T}] \rangle$;

- $B(\underline{T})$ is a non-empty conjunction of literals on a set of auxiliary predicates that can include atoms of the form $holds\_at(\ell, T')$, where $\ell[T']$ is a timed fluent literal, and the atom $time\_now(T'')$ for some variables $T', T''$;

- the conditions of the rule are constrained by the (possibly empty) temporal constraints $C[\underline{T}]$.

Any such rule again represents all of its ground instances under any total valuation of the variables in $\underline{T}$ that satisfies the constraints $C[\underline{T}]$. Each ground instance is named by the corresponding ground instance of $n(\tau, \underline{T})$. Intuitively, when the conditions of one such rule are satisfied at a time $now$ that grounds the variable $T''$ with the current time at which the capability is applied, then the goal in the head of the rule is sanctioned as one of the goals that the agent would possibly prefer to achieve at this time. The decision whether such a goal is indeed preferred would then depend on the high-level or strategy part $H$ of $KB_{GD}$, containing priority rules, as described in Section 3.3, between the rules in the lower-part or between other rules in $H$. These priority rules can also include temporal atoms of the form $holds\_at(\ell, T')$ and the atom $time\_now(T'')$ in their conditions.

To accommodate this form of rules we only need to extend our notion of incompatibility $I$ in $\mathcal{T}$ to be defined on conclusions $\langle \ell(\tau), TC[\tau, \underline{T}] \rangle$. To simplify the notation, in the remainder we often write $\langle \ell(\tau), TC \rangle$ instead of $\langle \ell(\tau), TC[\tau, \underline{T}] \rangle$.

The incompatibility $I$ can be defined in different ways. For example, a (relatively) weak notion of incompatibility is given as follows. Two pairs $\langle \ell_1(\tau_1), TC_1 \rangle$ and $\langle \ell_2(\tau_2), TC_2 \rangle$ are incompatible iff for every valuation $\sigma$ such that $TC_1$ and $TC_2$ are both satisfied, the ground instances of $\ell_1(\tau_1)\sigma$ and $\ell_2(\tau_2)\sigma$ are incompatible. A stronger notion would require that it is sufficient for only one such valuation $\sigma$ to exist that makes the corresponding ground literals incompatible.

Let us denote by $KB_{GD}^{now}$ the theory $KB_{GD} \cup \{time\_now(now)\}$, where $now$ is a time constant. Then, the goal decision capability, $\models_{GD}^{now}$, is defined directly in terms of the preference entailment, $\models_{pr}$, of LPP (see Section 3.3), as follows.

Given a state $S = \langle KB_0, \mathcal{F}, \mathcal{C}, \Sigma \rangle$,

$$S \models_{GD}^{now} \mathcal{G}s$$

where

$$\mathcal{G}s = \{G_1, G_2, \ldots, G_n\}, n \geq 0, G_i = \langle \ell_i(\tau_i), TC_i \rangle \text{ for all } i = 1, \ldots, n$$

iff $\mathcal{G}s$ is a maximal set such that

$$KB_{GD}^{now} \models_{pr} \langle \ell_1(\tau_1), TC_1 \rangle \wedge \ldots \wedge \langle \ell_n(\tau_n), TC_n \rangle.$$

This means that a new set of goals $\mathcal{G}s$ is generated that is currently (sceptically) preferred under the goal preference policy represented in $KB_{GD}$ and the current information in $KB_0$. Note that any two goals in $\mathcal{G}s$ are necessarily compatible with each other. There are two special cases where there are no sceptically preferred goals at the time *now*. The first one concerns the case where there are no goals that are currently sanctioned by the (lower-part) of $KB_{GD}$. When this is so $\models_{GD}^{now}$ returns an empty set of goals ($n = 0$). The second special case occurs when there are at least two goals which are each separately credulously preferred but these goals are incompatible which each other. Then $S \models_{GD}^{now} \bot$, where $\bot$ is used to indicate failure in identifying new goals to be pursued.

As an example, consider the San Vincenzo scenario where the *psa* agent needs to decide whether to return home or to recharge its battery. The agent's goals are categorised and assigned priority according to their category and possibly other factors. The $KB_{GD}$ expressing this is given as follows:

- The low-level part contains the rules:

$$
\begin{aligned}
n(rh, \tau_1) : \ & \langle return\_home(\tau_1), \{\tau_1 < T'\} \rangle \leftarrow \\
& holds\_at(finished\_work, T), \\
& holds\_at(\neg at\_home, T), \\
& time\_now(T), \\
& T' = T + 6 \\
n(rb, \tau_2) : \ & \langle recharge\_battery(\tau_2), \{\tau_2 < T'\} \rangle \leftarrow \\
& holds\_at(low\_battery, T), \\
& time\_now(T), \\
& T' = T + 2
\end{aligned}
$$

- The auxiliary part contains, in addition to $KB_{TR}$ and $KB_0$, the following rules that specify the category of each goal and the relative urgency between these categories:

$$
\begin{aligned}
& typeof(return\_home, required) \\
& typeof(recharge\_battery, operational) \\
& more\_urgent\_wrt\_type(operational, required)
\end{aligned}
$$

- The incompatibility part consists of

$$incompatible(return\_home(T), recharge\_battery(T))$$

  Namely, the two goals are pairwise incompatible, i.e. the agent can only do one of these goals at a time.

- The high-level part contains the following priority rule:

$$
\begin{aligned}
gd\_pref(X,Y) : n(X, \_) \prec n(Y, \_) \;\leftarrow\; & typeof(X, XT), \\
& typeof(Y, YT), \\
& more\_urgent\_wrt\_type(XT, YT).
\end{aligned}
$$

Then, for $now = 1$ and current state $S = \langle KB_0, \mathcal{F}, \mathcal{C}, \Sigma \rangle$ such that finished work and away from home both hold (by temporal reasoning) at time $now$, we have that

$$S \models_{GD}^{now} \{\langle return\_home(\tau_1), \{\tau_1 < 7\}\rangle\}.$$

Suppose instead that $KB_0$ contains $observed(low\_battery, 1)$. Then, using the weak notion of incompatibility, requiring that

   for every $\sigma$ such that $\sigma \models_{cs} \{\tau_1 < 7, \tau_2 < 3\}$

   it holds that $incompatible(return\_home(\tau_1)\sigma, recharge\_battery(\tau_2)\sigma)$

we have:

$$S \models_{GD}^{now} \{\langle return\_home(\tau_1), \{\tau_1 < 7\}\rangle, \langle recharge\_battery(\tau_2), \{\tau_2 < 3\}\rangle\}.$$

Indeed, for $\sigma = \{\tau_1 = 3, \tau_2 = 2\}$, $incompatible(return\_home(3), recharge\_battery(2))$ does not hold. However, using the stronger notion of incompatibility, requiring that

   there exists $\sigma$ such that $\sigma \models_{cs} \{\tau_1 < 7, \tau_2 < 3\}$

   it holds that $incompatible(return\_home(\tau_1)\sigma, recharge\_battery(\tau_2)\sigma)$

we have:

$$S \models_{GD}^{now} \{\langle recharge\_battery(\tau_2), \{\tau_2 < 3\}\rangle\}.$$

Suppose now that $KB_{GD}$ contains a second operational goal $\langle replace\_part(\tau_3), \{\tau_3 < 5\}\rangle$ that is also sanctioned by a rule in its lower part at time $now = 1$. Then under the stronger form of incompatibility the goal decision capability at $now = 1$ will return $\bot$ as both these operational goals are credulously preferred but none of them is sceptically preferred.

## 6. Physical Capabilities

In addition to the reasoning capabilities we have defined so far, an agent is equipped with physical capabilities that allow it to experience the world in which it is situated; this world consists of other agents and/or objects that provide an environment for the agents in which to interact and communicate.

We identify two types of physical capabilities: *sensing* and *actuating*. In representing these capabilities we abstract away from the sensors and the actuators that an agent would typically rely upon to access and affect the environment. We will also assume that these sensors and actuators are part of the agent's body, which we classify as an implementation issue (Stathis et al., 2004).

The physical sensing capability models the way an agent interacts with its external environment in order to inspect it, e.g. to find out whether or not some fluent holds at a given time. On the other hand, the physical actuating capability models the way an agent interacts with its external environment in order to affect it, by physically executing its actions.

We represent the sensing physical capability of an agent as a function of the form:

$$\texttt{sensing}(L, t) = L'$$

where:

- $L$ is a (possibly empty) set of

  - fluent literals $f$,
  - terms of the form $c : a$ (meaning that agent $c$ has performed action $a$),

  all to be sensed at a concrete time $t$, and

- $L'$ is a (possibly empty) set of elements $s'$ such that

  - $s'$ is a term $f : v$, $f$ being a fluent and $v \in \{true, false\}$, meaning that fluent $f$ has been observed to have value $v$ (namely to be $true$ or to be $false$) at time $t$, or
  - $s'$ is a term of the form $c : a[t']$, $c$ being an agent name and $a$ being an action, meaning that agent $c$ has performed action $a$ at time $t'$.

Note that physical sensing requires the time-stamp $t$ to specify the time at which it is applied within transitions. Note also that, given a non-empty set $L$, $\texttt{sensing}(L, t)$ may be partial, e.g. for some fluent $f \in L$, it can be that neither $f : true \in L'$, nor $f : false \in L'$.

Similarly, we represent the physical actuating capability as a function

$$\texttt{actuating}(As, t) = As'$$

where:

- $As$ is a set of action literals $\{a_1, \cdots, a_n\}$, $n > 0$, that the agent instructs the body to actuate at time $t$;

- $As' \subseteq As$ is the subset of actions that the body has actually managed to perform.

The meaning of an action $a$ belonging to $As$ and not belonging to $As'$ is that the physical actuators of the agent's body were not able to perform $a$ in the current situation. It is worth pointing out that if an action $a$ belongs to $As'$ it does not necessarily mean that the effects of $a$ have successfully been reached. Indeed, some of the preconditions of the executed action (i) may have been wrongly believed by the agent to be true at execution time (as other agents may have interfered with them) or (ii) the agent may have been unaware of these preconditions. For example, after having confirmed availability, the agent may have booked a hotel by sending an e-mail, but (i) some other agent has booked the last available room in the meanwhile, or (ii) the agent did not provide a credit card number to secure the booking. In other words, the beliefs of the agent (as held in $KB_0$) may be incorrect and/or incomplete.

In Section 7 and Section 8 below, we will see that AOI (Active Observation Introduction) can be used to check effects of actions (identified by the $f_{ES}$ effect selection operator, in turn using the $\models_{eff}$ reasoning capability) after actions have been executed. Moreover, SI (Sensing Introduction) can be used to check preconditions of actions (identified by the $f_{PS}$ precondition selection operator, in turn using the $\models_{pre}$ reasoning capability) just before they are executed, to make sure that the actions are indeed executable. Overall, the following cases may occur:

- an action belongs to $As'$ because it was executed and

  - its preconditions held at the time of execution and its effects hold in the environment after execution;

  - its preconditions were wrongly believed to hold at the time of execution (because the agent has partial knowledge of the environment or its $KB_{plan}$ is incorrect) and as a consequence its effects do not hold after execution;

  - its preconditions were known not to hold at the time of execution (e.g. because the agent observed only after having planned that they did not hold, but had no time to -replan) and as a consequence its effects do not hold after execution;

- an action belongs to $As \setminus As'$ because it was not executed (the body could not execute it).

The `actuating` physical capability does not check preconditions/effects: this is left to other capabilities called within transitions before and/or after the transition invoking `actuating`, as we will show below. As before, the way the body will carry out the actions is an implementation issue (Stathis et al., 2004).

## 7. Transitions

The KGP model relies upon the state transitions GI, PI, RE, SI, POI, AOI, AE, SR, defined below using the following representation

$$(\mathbf{T}) \qquad \frac{\langle KB_0, \mathcal{F}, \mathcal{C}, \Sigma \rangle \qquad X}{\langle KB_0', \mathcal{F}', \mathcal{C}', \Sigma' \rangle} \, now$$

where **T** is the name of the transition, $\langle KB_0, \mathcal{F}, \mathcal{C}, \Sigma \rangle$ is the agent's state before the transition is applied, $X$ is the input for the transition, *now* is the time of application of the transition, $\langle KB_0', \mathcal{F}', \mathcal{C}', \Sigma' \rangle$ is the revised state, resulting from the application of the transition **T** with input $X$ at time *now* in state $\langle KB_0, \mathcal{F}, \mathcal{C}, \Sigma \rangle$. Please note that most transitions only modify some of the components of the state. Also, for some transitions (namely GI, RE, POI, SR) the input $X$ is always empty and will be omitted. For the other transitions (namely PI, SI, AOI, AE) the input is always non-empty (see Section 9) and is selected by an appropriate selection operator (see Section 8).

Below we define each transition formally, by defining $\langle KB_0', \mathcal{F}', \mathcal{C}', \Sigma' \rangle$. Note that we assume that each transition takes care of possible renaming of time variables in the output of capabilities (if a capability is used by the transition), in order to guarantee that each goal/action in the forest is univocally identified by its time variable.

## 7.1 Goal Introduction

This transition takes empty input. It calls the Goal Decision capability to determine the new (non-reactive) top-level goals of the agent. If this capability returns a set of goals, this means that the circumstances have now possibly changed the preferred top-level goals of the agent and the transition will reflect this by changing the forest in the new state to consist of one tree for each new (non-reactive) goal. On the other hand, if the Goal Decision capability does not return any (non-reactive) goals (namely it returns $\bot$) the state is left unchanged, as, although the goals in the current state are no longer sceptically preferred they may still be credulously preferred and, since there are no others to replace them, the agent will carry on with its current plans to achieve them.

$$(\mathbf{GI}) \qquad \frac{\langle KB_0, \mathcal{F}, \mathcal{C}, \Sigma \rangle}{\langle KB_0, \mathcal{F}', \mathcal{C}', \Sigma \rangle} \; now$$

where, given that $S = \langle KB_0, \mathcal{F}, \mathcal{C}, \Sigma \rangle$

(i) If $S \models_{GD}^{now} \bot$, then

     − $\mathcal{F}' = \mathcal{F}$

     − $\mathcal{C}' = \mathcal{C}$

(ii) otherwise, if $S \models_{GD}^{now} Gs$ and $Gs \neq \bot$, then

     − $\mathcal{F}'$ is defined as follows:

         ∗ $nr(\mathcal{F}') = \{ \mathcal{T}_{g[\tau]} \mid \langle g[\tau], \_ \rangle \in Gs \}$ where $\mathcal{T}_{g[\tau]}$ is a tree consisting solely of the root $g[\tau]$

         ∗ $r(\mathcal{F}') = \{\}$

     − $\mathcal{C}' = \{ TC \mid \langle \_, TC \rangle \in Gs \}$

This transition drops (top-level) goals that have become "semantically" irrelevant (due to changed circumstances of the agent or changes in its environment), and replaces them with new relevant goals. We will see, in Section 7.8, that goals can also be dropped because of the book-keeping activities of the State Revision (SR) transition, but that transition can never add to the set of goals.

Note that, as GI will replace the whole forest in the old state by a new forest, it is possible that the agent looses valuable information that it has in achieving its goals, when one of the new preferred goals of the agent is the same as (or equivalent to) a current goal. This effect though can be minimized by calling (in the cycle theory) the GI transition only at certain times, e.g. after the current goals have been achieved or timed-out. Alternatively, the earlier formalisation of the GI transition could be modified so that, in case (ii), for all goals in $Gs$ that already occur (modulo their temporal variables and associated temporal constraints) as roots of (non-reactive) trees in $\mathcal{F}$, these trees are kept in $\mathcal{F}'$. A simple way to characterise (some of) these goals is as follows. Let

$$
\begin{aligned}
Xs = \{\langle g[\tau], TC, \tau = \tau'\rangle \mid \quad &\langle g[\tau], TC\rangle \in Gs, \\
&g[\tau'] \in Roots^{nr}(\mathcal{F}) \text{ and} \\
&\models_{cs} \mathcal{C} \text{ iff } \models_{cs} (\mathcal{C} \cup TC \cup \{\tau = \tau'\}))\} \\
Gs' = \{\langle g[\tau], TC\rangle \mid \quad &\langle g[\tau], TC, \tau = \tau'\rangle \in Xs\}
\end{aligned}
$$

The new constraints on goals in $Gs'$ are equivalent to the old constraints in $\mathcal{C}$. For example, $Gs$ may contain

$\quad G = \langle have\_ticket(madrid, denver, \tau_2), \{\tau_2 < 12\}\rangle$

$\quad$ with $have\_ticket(madrid, denver, \tau_1) \in Roots^{nr}(\mathcal{F})$ and $\mathcal{C} = \{\tau_1 < 12\}$.

Then, $G$ definitely belongs to $Gs'$. Let

$$
new\mathcal{C} = \bigcup_{\langle \_, TC, \tau = \tau'\rangle \in Xs} TC \cup \{\tau = \tau'\}.
$$

Case (ii) can be redefined as follows, using these definitions of $Xs, Gs'$ and $new\mathcal{C}$:

(ii′) otherwise, if $S \models_{GD}^{now} Gs$ and $Gs \neq \bot$, then, if it is not the case that $\models_{cs} \mathcal{C} \cup new\mathcal{C}$, then $\mathcal{F}'$ and $\mathcal{C}'$ are defined as in the earlier case (ii), otherwise (if $\models_{cs} \mathcal{C} \cup new\mathcal{C}$):

   – $\mathcal{F}'$ is defined as follows:
   
   * $nr(\mathcal{F}') = \{\mathcal{T}_{g[\tau]} \mid \langle g[\tau], \_\rangle \in Gs \setminus Gs'\} \cup \mathcal{F}(Xs)$
   where $\mathcal{T}_{g[\tau]}$ is a tree consisting solely of the root $g[\tau]$ and $\mathcal{F}(Xs)$ is the set of all trees in $\mathcal{F}$ with roots goals of the form $g[\tau']$ such that $\langle g[\tau], \_, \tau = \tau'\rangle \in Xs$
   * $r(\mathcal{F}') = \{\}$
   – $\mathcal{C}' = \mathcal{C} \cup \{TC \mid \langle \_, TC\rangle \in Gs \setminus Gs'\} \cup new\mathcal{C}$.

Note that we keep all temporal constraints in the state, prior to the application of GI, but we force all variables of new goals that remain in the state after GI to be rewritten using the old identifiers of the goals.

## 7.2 Reactivity

This transition takes empty input. It calls the Reactivity capability in order to determine the new top-level reactive goals in the state (if any), leaving the non-reactive part unchanged. If no new reactive goals exist, the reactive part of the new state will be empty.

$$(\mathbf{RE}) \qquad \frac{\langle KB_0, \mathcal{F}, \mathcal{C}, \Sigma \rangle}{\langle KB_0, \mathcal{F}', \mathcal{C}', \Sigma \rangle} \, now$$

where, given that $S = \langle KB_0, \mathcal{F}, \mathcal{C}, \Sigma \rangle$:

(i) If $S \models_{react}^{now} \bot$, then

- $\mathcal{F}'$ is defined as follows:
    - $r(\mathcal{F}') = \{\}$
    - $nr(\mathcal{F}') = nr(\mathcal{F})$
- $\mathcal{C}' = \mathcal{C}$

(ii) otherwise, if $S \models_{react}^{now} (\mathcal{X}s, TC)$, then

- $\mathcal{F}'$ is defined as follows:
    - $nr(\mathcal{F}') = nr(\mathcal{F})$
    - $r(\mathcal{F}') = \{\mathcal{T}_{x[\tau]} \mid x[\tau] \in \mathcal{X}s\}$
      where $\mathcal{T}_{x[\tau]}$ is a tree consisting solely of the root $x[\tau]$
- $\mathcal{C}' = \mathcal{C} \cup TC$

Note that there is an asymmetry between case (ii) of GI and case (ii) of RE, as GI eliminates all reactive goals in this case, whereas RE leaves all non-reactive goals unchanged. Indeed, reactive goals may be due to the choice of specific non-reactive goals, so when the latter change the former need to be re-evaluated. Instead, non-reactive goals are not affected by newly acquired reactive goals (that are the outcome of enforcing reactive rules).

Note also that in case (ii), similarly to GI, as RE replaces the whole (reactive) forest in the old state by a new (reactive) forest, it is possible that the agent loses valuable information that it has in achieving its reactive goals, when one of the new reactive goals is the same as (or equivalent to) a current goal. A variant of case (ii) for RE, mirroring the variant given earlier for GI and using $\models_{cs}$ as well, can be defined to avoid this problem.

## 7.3 Plan Introduction

This transition takes as input a non-executable goal in the state (that has been selected by the *goal selection operator*, see Section 8) and produces a new state by calling the agent's Planning capability, if the selected goal is a mental goal, or by simply introducing a new sensing action, if the goal is a sensing goal.

$$(\mathbf{PI}) \qquad \frac{\langle KB_0, \mathcal{F}, \mathcal{C}, \Sigma \rangle \quad G}{\langle KB_0, \mathcal{F}', \mathcal{C}', \Sigma \rangle} \, now$$

where $G$ is the input goal (selected for planning in some tree $\mathcal{T}$ in $\mathcal{F}$, and thus a leaf, see Section 8) and

$\mathcal{F}' = (\mathcal{F} \setminus \{\mathcal{T} \mid G \text{ is a leaf in } \mathcal{T}\}) \cup New$

$\mathcal{C}' = \mathcal{C} \cup TC$

where $New$ and $TC$ are obtained as follows, $S$ being $\langle KB_0, \mathcal{F}, \mathcal{C}, \Sigma \rangle$.

(i) if $G$ is a mental goal: let $S, G \models_{plan}^{now} P$. Then,

  – either $P = \bot$ and
  $New = \{\mathcal{T}\}$ and $TC = \{\}$,

  – or $P = (\mathcal{X}s, TC)$ and
  $New = \{\mathcal{T}'\}$ where $\mathcal{T}'$ is obtained from $\mathcal{T}$ by adding each element of $\mathcal{X}s$ as a child of $G$.

(ii) if $G = \ell[\tau]$ is a sensing goal, and a child of a goal $G'$ in $\mathcal{T}$:

  $New = \{\mathcal{T}'\}$ where $\mathcal{T}'$ is $\mathcal{T}$ with (a node labelled by) $sense(\ell, \tau')$ as a new child of $G'$ (here $\tau'$ is a new time variable) and

  $TC = \{\tau' \le \tau\}$.

(iii) if $G = \ell[\tau]$ is a sensing goal, and the root of $\mathcal{T}$:

  $New = \{\mathcal{T}, \mathcal{T}'\}$ where $\mathcal{T}'$ is a tree consisting solely of the root (labelled by) $sense(\ell, \tau')$ (here $\tau'$ is a new time variable) and

  $TC = \{\tau' \le \tau\}$.

## 7.4 Sensing Introduction

This transition takes as input a set of fluent literals that are preconditions of some actions in the state and produces a new state by adding sensing actions as leaves in (appropriate) trees in its forest component. Note that, when SI is invoked, these input fluent literals are selected by the *precondition selection operator*, and are chosen amongst preconditions of actions that are not already known to be true (see Section 8).

$$(\mathbf{SI}) \quad \frac{\langle KB_0, \mathcal{F}, \mathcal{C}, \Sigma \rangle \quad SPs}{\langle KB_0, \mathcal{F}', \mathcal{C}', \Sigma \rangle} now$$

with $SPs$ a non-empty set of preconditions of actions (in the form of pairs "precondition, action") in some trees in $\mathcal{F}$, where, given that:

- $New = \{\langle \ell[\tau], A, sense(\ell, \tau') \rangle \mid \langle \ell[\tau], A \rangle \in SPs \text{ and } \tau' \text{ is a fresh variable}\}$

- $addSibling(\mathcal{T}, A, SA)$ denotes the tree obtained by adding all elements in $SA$ as new siblings of $A$ to the tree $\mathcal{T}$ such that $leaf(A, \mathcal{T})$

then

$$\begin{aligned}
\mathcal{F}' \quad = \quad & \mathcal{F} \setminus \{\mathcal{T} \mid leaf(A, \mathcal{T}) \text{ and } \langle \ell[\tau], A \rangle \in SPs\} \\
& \cup \{addSibling(\mathcal{T}, A, SA) \mid leaf(A, \mathcal{T}) \text{ and } \\
& \quad SA = \{sense(\ell, \tau') \mid \langle \ell[\tau], A, sense(\ell[\tau']) \rangle \in New\}\}
\end{aligned}$$

$$\begin{aligned}
\mathcal{C}' \quad = \quad & \mathcal{C} \cup \{\tau' < \tau \mid \langle \ell[\tau], \_, sense(\ell[\tau']) \rangle \in New\}
\end{aligned}$$

Basically, for each fluent literal selected by the precondition selection operator as a precondition of an action $A$, a new sensing action is added as a sibling of $A$, and the constraint expressing that this sensing action must be performed before $A$ is added to the current set of temporal constraints.

### 7.5 Passive Observation Introduction

This transition updates $KB_0$ by adding new observed facts reflecting changes in the environment. These observations are not deliberately made by the agent, rather, they are "forced" upon the agent by the environment. These observations may be properties in the form of positive or negative fluents (for example that the battery is running out) or actions performed by other agents (for example messages addressed to the agent).

$$(\textbf{POI}) \qquad \frac{\langle KB_0, \mathcal{F}, \mathcal{C}, \Sigma \rangle}{\langle KB_0', \mathcal{F}, \mathcal{C}, \Sigma \rangle} \, now$$

where, if $\texttt{sensing}(\emptyset, now) = L$, then

$$
\begin{aligned}
KB_0' \;=\; & KB_0 \,\cup \\
& \{observed(f, now) \mid f : true \in L\} \,\cup \\
& \{observed(\neg f, now) \mid f : false \in L\} \,\cup \\
& \{observed(c, a[t], now) \mid c : a[t] \in L\}.
\end{aligned}
$$

### 7.6 Active Observation Introduction

This transition updates $KB_0$ by adding new facts deliberately observed by the agent, which seeks to establish whether or not some given fluents hold at a given time. These fluents are selected by the *effect selection operator* (see Section 8) and given as input to the transition. Whereas POI is not "decided" by the agent (the agent is "interrupted" and forced an observation by the environment), AOI is deliberate. Moreover, POI may observe fluents and actions, whereas AOI only considers fluents (that are effects of actions executed by the agent, as we will see in Section 8 and in Section 9).

$$(\textbf{AOI}) \qquad \frac{\langle KB_0, \mathcal{F}, \mathcal{C}, \Sigma \rangle \quad SFs}{\langle KB_0', \mathcal{F}, \mathcal{C}, \Sigma \rangle} \, now$$

where $SFs = \{f_1, \ldots, f_n\}$, $n > 0$, is a set of fluents selected for being actively sensed (by the effect selection operator), and, if $\texttt{sensing}(SFs, now) = L$, then

$$
\begin{aligned}
KB_0' \;=\; & KB_0 \,\cup \\
& \{observed(f, now) \mid f : true \in L\} \,\cup \\
& \{observed(\neg f, now) \mid f : false \in L\}.
\end{aligned}
$$

### 7.7 Action Execution

This transition updates $KB_0$, recording the execution of actions by the agent. The actions to be executed are selected by the *action selection operator* (see Section 8) prior to the transition, and given as input to the transition.

$$(\mathbf{AE}) \qquad \frac{\langle KB_0, \mathcal{F}, \mathcal{C}, \Sigma \rangle \qquad SAs}{\langle KB_0', \mathcal{F}, \mathcal{C}, \Sigma' \rangle} \, now$$

where $SAs$ is a non-empty set of actions selected for execution (by the action selection operator), and

- let $A$ be the subset of all non-sensing actions in $SAs$ and $S$ be the subset of all sensing actions in $SAs$;

- let $\mathtt{sensing}(S', now) = L'$, where $S' = \{f \mid sense(f, \tau) \in S\}$

- let $\mathtt{sensing}(S'', now) = L''$, where $S'' = \{c : a \mid sense(c : a, \tau) \in S\}$

- let $\mathtt{actuating}(A', now) = A''$, where $A' = \{a \mid a[\tau] \in A\}$.

Then:

$$
\begin{aligned}
KB_0' \;=\; & KB_0 \cup \\
& \{executed(a, now) \mid a \in A''\} \;\cup \\
& \{observed(f, now) \mid f : true \in L'\} \;\cup \\
& \{observed(\neg f, now) \mid f : false \in L'\} \\
& \{observed(c, a[t], now) \mid c : a[t] \in L'' \text{ and} \\
& \qquad\qquad \exists \sigma \text{ such that } \sigma \models_{cs} \mathcal{C} \wedge \tau = t \text{ where } sense(c : a, \tau) \in S\}
\end{aligned}
$$

and

$$
\begin{aligned}
\Sigma' \;=\; & \Sigma \cup \{\tau = now \mid a[\tau] \in SAs \wedge a \in A''\} \cup \\
& \{\tau = now \mid sense(f, \tau) \in SAs \wedge (f : \_) \in L'\} \cup \\
& \{\tau = t \mid c : a[t] \in L'' \text{ and } \exists \sigma \text{ such that } \sigma \models_{cs} \mathcal{C} \wedge \tau = t \text{ where } sense(c : a, \tau) \in S\}.
\end{aligned}
$$

## 7.8 State Revision

The SR transition revises a state by removing all timed-out goals and actions and all goals and actions that have become obsolete because one of their ancestors is already believed to have been achieved. We will make use of the following terminology.

**Notation 7.1** Given a state $S$, a timed fluent literal $\ell[\tau]$, a timed fluent literal or action operator $x[\tau]$, and a time-point $now$:

- $achieved(S, \ell[\tau], now)$ stands for

  there exists a total valuation $\sigma$ such that $S, \sigma \models_{cs} \tau \leq now$ and $S \models_{TR} \ell[\tau]\sigma$

- $timed\_out(S, x[\tau], now)$ stands for

  there exists no total valuation $\sigma$ such that $S, \sigma \models_{cs} \tau > now$.

Then, the specification of the transition is as follows.

$$(\textbf{SR}) \qquad \frac{\langle KB_0, \mathcal{F}, \mathcal{C}, \Sigma \rangle}{\langle KB_0, \mathcal{F}', \mathcal{C}, \Sigma \rangle} \, now$$

where $\mathcal{F}'$ is the set of all trees in $\mathcal{F}$ pruned so that $nodes(\mathcal{F}')$ is the biggest subset of $nodes(\mathcal{F})$ consisting of all goals/actions $x[\tau]$ in some tree $\mathcal{T}$ in $\mathcal{F}$ such that (here $S = \langle KB_0, \mathcal{F}, \mathcal{C}, \Sigma \rangle$):

(i) $\neg timed\_out(S, x[\tau], now)$, and

(ii) if $x$ is an action operator, it is not the case that $executed(x, t) \in KB_0$ and $(\tau = t) \in \Sigma$, and

(iii) if $x$ is a fluent literal, $\neg achieved(S, x[\tau], now)$, and

(iv) for every $y[\tau'] \in siblings(x[\tau], \mathcal{F})$

    – either $y[\tau'] \in siblings(x[\tau], \mathcal{F}')$,

    – or $y[\tau'] \notin siblings(x[\tau], \mathcal{F}')$ and

        ∗ if $y$ is a fluent literal then $achieved(S, y[\tau'], now)$,

        ∗ if $y$ is an action literal then $executed(y, t) \in KB_0$ and $\tau' = t \in \Sigma$,

    and

(v) if $x$ is a *sensing* action operator, $x[\tau] = sense(\ell, \tau)$, then

    – either there exists $a[\tau'] \in siblings(x[\tau], \mathcal{F}')$ such that $\ell$ is a precondition of $a$ (i.e. $S, a[\tau'] \models_{pre} Cs$ and $\ell[\tau'] \in Cs$) and $\tau < \tau' \in \mathcal{C}$,

    – or there exists $\ell[\tau'] \in siblings(x[\tau], \mathcal{F}')$ such that $\ell$ is a sensing fluent and $\tau < \tau' \in \mathcal{C}$, and

(vi) $x[\tau]$ is a top-level goal or $parent(x[\tau], \mathcal{F}) = P$ and $P \in nodes(\mathcal{F}')$.

All conditions above specify what SR keeps in the trees in the forest in the state. Intuitively, these conditions may be understood in terms of what they prevent from remaining in such trees:

- condition (i) removes *timed-out* goals and actions,

- condition (ii) removes actions that have already been executed,

- condition (iii) removes goals that are already achieved,

- condition (iv) removes goals and actions whose siblings are already timed out and thus deleted, by condition (i),

- condition (v) removes sensing actions for preconditions of actions that have been deleted and for sensing goals that have been deleted,

- condition (vi) recursively removes actions and goals whose ancestors have been removed.

The following example illustrates how SR is used to provide adjustment of the agent's goals and plans in the light of newly acquired information.

### 7.9 Setting 3

The agent *psa* has the goal to have a museum ticket for some (state-run) museum that the businessman wants to visit, and a plan to buy the ticket. But before executing the plan *psa* observes that it is the European Heritage day (*ehd* for short), via an appropriate "message" from another agent *mus* (representing the museum), stating that all state-run museums in Europe give out free tickets to anybody walking in on that day. Then, the *psa*'s goal is already achieved and both goal and plan are deleted from its state.

Let the agent's initial state be $\langle KB_0, \mathcal{F}, \mathcal{C}, \Sigma \rangle$ with:

$$
\begin{aligned}
\Sigma &= \{\ \} = KB_0 \\
\mathcal{F} &= \{\mathcal{T}\} \\
\mathcal{C} &= \{\tau_1 \leq 10, \tau_2 = \tau_3, \tau_3 < \tau_1\}
\end{aligned}
$$

where $\mathcal{T}$ consists of a top-level goal $g_1 = have(ticket, \tau_1)$, with two children,

$g_2 = have\_money(\tau_2)$ and $a_1 = buy(ticket, \tau_3)$, [15]

and further assuming that $KB_{TR}$ contains

$$
\begin{aligned}
&initiates(ehd, T, have(ticket)) \\
&initiates(buy(O), T, have(O)) \\
&precondition(buy(O), have\_money).
\end{aligned}
$$

The remaining knowledge bases do not play any useful role for the purposes of this example, and can therefore be considered to be empty. The "message" from the museum agent *mus* is added to $KB_0$ via POI, e.g. at time 6, in the following form:

$$observed(mus, ehd(5), 6)$$

i.e. at time 6 it is observed that at time 5 *mus* has announced that all state-run museums in Europe are free on that day. Then, via SR, at time 8 say, $g_1$, $g_2$ and $a_1$ are eliminated from $\mathcal{F}$, as $g_1$ is already achieved.

## 8. Selection Operators

The KGP model relies upon selection operators:

- $f_{GS}$ (goal selection, used to provide input to the PI transition);

- $f_{PS}$ (precondition selection, used to provide input to the SI transition);

- $f_{ES}$ (effect selection, used to provide input to the AOI transition);

- $f_{AS}$ (action selection, used to provide input to the AE transition).

---

15. $g_1$ and $a_1$ can be reactive or not, as this does not matter for this example.

Selection operators are defined in terms of (some of the) capabilities (namely Temporal Reasoning, Identification of Preconditions and Effects and Constraint Solving).

At a high-level of description, the selection operators can all be seen as returning the set of all items from a given initial set that satisfy a certain number of conditions. For example, given a state $\langle KB_0, \mathcal{F}, \mathcal{C}, \Sigma \rangle$, the goal selection operator returns the set of all non-executable goals in trees in $\mathcal{F}$ that satisfy some conditions; the precondition selection operator returns the set of all pairs, each consisting of (i) a timed fluent literal which is a precondition of some action in some tree in $\mathcal{F}$ and (ii) that action, satisfying some conditions; the effect selection operator returns the set of all fluent literals which are effects of actions already executed (as recorded in $KB_0$) that satisfy some conditions; the action selection operator returns the set of all actions in trees in $\mathcal{F}$ that satisfy some conditions.

The selection operators are formally defined below.

## 8.1 Goal Selection

Informally, the set of conditions for the goal selection operator is as follows. Given a state $S = \langle KB_0, \mathcal{F}, \mathcal{C}, \Sigma \rangle$ and a time-point $t$, the set of goals selected by $f_{GS}$ is a singleton set consisting of a non-executable goal $G$ in some tree in $\mathcal{F}$ such that at time $t$:

1. $G$ is not timed out,

2. no ancestor of $G$ is timed out,

3. no child of any ancestor of $G$ is timed out,

4. neither $G$, nor any ancestor of $G$ in any tree in $\mathcal{F}$ is already achieved.

5. $G$ is a leaf

Intuitively, condition 1 ensures that $G$ is not already timed-out, conditions 2-3 impose that $G$ belongs to a "still feasible" plan for some top-level goal in $\mathcal{F}$, and condition 4 makes sure that considering $G$ is not wasteful.

Note that, as already mentioned in Section 5.1.3, for simplicity we select a single goal. Formally, given a state $S = \langle KB_0, \mathcal{F}, \mathcal{C}, \Sigma \rangle$ and a time-point $t$, let $\mathcal{G}(S, t)$ be the set of all non-executable goals $\ell[\tau] \in nodes(\mathcal{F})$ such that:

1. $\neg timed\_out(S, \ell[\tau], t)$

2. $\neg timed\_out(S, G, t)$ for each $G \in ancestors(\ell[\tau], \mathcal{F})$,

3. $\neg timed\_out(S, X, t)$ for each $X \in nodes(\mathcal{F})$ such that $X$ is the child of some $P \in ancestors(\ell[\tau], \mathcal{F})$

4. $\neg achieved(S, G, t)$ for each $G \in \{\ell[\tau]\} \cup ancestors(\ell[\tau], \mathcal{F})$

5. $leaf(G, \mathcal{F})$

Then, if $\mathcal{G}(S, t) \neq \{\}$:

$$f_{GS}(S, t) = \{G\} \text{ for some } G \in \mathcal{G}(S, t).$$

Otherwise, $f_{GS}(S, t) = \{\}$.

## 8.2 Effect Selection

Informally, the set of conditions for the effect selection operator is as follows. Given a state $S = \langle KB_0, \mathcal{F}, \mathcal{C}, \Sigma \rangle$ and a time-point $t$, $f_{ES}$ selects all fluents $f$ such that $f$ or $\neg f$ is one of the effects of some action $a[\tau]$ that has "recently" been executed.

Note that such $f$ (or $\neg f$) may not occur in $\mathcal{F}$ but could be some other (observable) effect of the executed action, which is not necessarily the same as the goal that the action contributes to achieving. For example, in order to check whether an internet connection is available, the agent may want to observe that it can access a skype network even though it is really interested in opening a browser (as it needs a browser in order to perform a booking online).

Formally, given a state $S = \langle KB_0, \mathcal{F}, \mathcal{C}, \Sigma \rangle$ and a time-point $now$, the set of all (timed) fluents selected by $f_{ES}$ is the set of all (timed) fluents $f[\tau]$ such that there is an action operator $a$ with

1. $executed(a, t') \in KB_0$, $t' = \tau \in \Sigma$ and $now - \epsilon < t' < now$, where $\epsilon$ is a sufficiently small number (that is left as a parameter here), and

2. $S, a[\tau] \models_{eff} \ell$, where $\ell = f$ or $\ell = \neg f$.

## 8.3 Action Selection

Informally, the set of conditions for the action selection operator is as follows. Given a state $S = \langle KB_0, \mathcal{F}, \mathcal{C}, \Sigma \rangle$ and a time-point $t$, the set of all actions selected by $f_{AS}$ is defined as follows. Let $\mathcal{X}(S, t)$ be the set of all actions $A$ in trees in $\mathcal{F}$ such that:

1. $A$ can be executed,

2. no ancestor of $A$ is timed out,

3. no child of any ancestor of $A$ is timed out,

4. no ancestor of $A$ is already satisfied,

5. no precondition of $A$ is known to be false,

6. $A$ has not already been executed.

Then $f_{AS}(S, t) \subseteq \mathcal{X}(S, t)$ such that all actions in $f_{AS}(S, t)$ are executable concurrently at $t$.

Intuitively, conditions 2-4 impose that $A$ belongs to a "still feasible" plan for some top-level goals in $\mathcal{F}$. Note that condition 1 in the definition of $\mathcal{X}(S, t)$ is logically redundant, as it is also re-imposed by definition of $f_{AS}(S, t)$. However, this condition serves as a first filter and is thus useful in practice.

Formally, given a state $S = \langle KB_0, \mathcal{F}, \mathcal{C}, \Sigma \rangle$, and a time-point $t$, the set of all actions selected by $f_{AS}$ is defined as follows. Let $\mathcal{X}(S, t)$ be the set of all actions $a[\tau]$ occurring as leaves of some trees in $\mathcal{F}$ such that:

1. there exists a total valuation $\sigma$ such that $S, \sigma \models_{cs} \tau = t$, and

2. $\neg timed\_out(S, G, t)$ for each $G \in ancestors(a[\tau], \mathcal{F})$, and

3. $\neg timed\_out(S, X, t)$ for each $X \in children(G, \mathcal{F})$ and $G \in ancestors(a[\tau], \mathcal{F})$, and

4. $\neg achieved(S, G, t)$ for each $G \in ancestors(a[\tau], \mathcal{F})$, and

5. let $S, a[\tau] \models_{pre} Cs$ and $Cs = \ell_1[\tau] \wedge \ldots \wedge \ell_n[\tau]$;

   if $n > 0$, then for no $i = 1, \ldots, n$ there exists a total valuation $\sigma$ such that $S, \sigma \models_{cs} \tau = t$ and $S \models_{TR} \overline{\ell_i}[\tau]\sigma$, and

6. there exists no $t'$ such that $\tau = t' \in \Sigma$ and $executed(a, t') \in KB_0$.

The formalisation of condition 6 allows for other instances of action $A$ to have been executed. Then,

$$f_{AS}(S, t) = \{a_1[\tau_1], \ldots, a_m[\tau_m]\} \subseteq \mathcal{X}(S, t)$$

(where $m \geq 0$), such that there exists a total valuation $\sigma$ for the variables in $\mathcal{C}$ such that $S, \sigma \models_{cs} \tau_1 = t \wedge \ldots \wedge \tau_m = t$.

Note that the definition of the action selection operator can be extended to take into account a notion of urgency with respect to the temporal constraints. However, such an extension is beyond the scope of this work.

## 8.4 Precondition Selection

Informally, the set of conditions for the precondition selection operator is as follows. Given a state $S = \langle KB_0, \mathcal{F}, \mathcal{C}, \Sigma \rangle$ and a time-point $t$, the set of preconditions (of actions in $\mathcal{F}$) selected by $f_{PS}$ is the set of all pairs $\langle C, A \rangle$ of (timed) preconditions $C$ and actions $A \in nodes(\mathcal{F})$ such that:

1. $C$ is a precondition of $A$ and

2. $C$ is not known to be true in $S$ at $t$, and

3. $A$ is one of the actions that could be selected for execution if $f_{AS}$ would be called at the current time.

The reason why this selection operator returns pairs, rather then simply preconditions, is that the transition SI, which makes use of the outputs of this selection operator, needs to know the actions associated with the preconditions. This is because SI introduces sensing actions for each precondition returned and has to place these sensing actions as siblings of the associated actions in $\mathcal{F}$, as seen in Section 7.4.

Formally, given a state $S = \langle KB_0, \mathcal{F}, \mathcal{C}, \Sigma \rangle$ and a time-point $t$, the set of all preconditions of actions selected by $f_{PS}$ is the set of all pairs $\langle C, A \rangle$ of (timed) preconditions $C$ and actions $A \in nodes(\mathcal{F})$ such that:

1. $A = a[\tau]$, and $S, a[\tau] \models_{pre} Cs$ and $C$ is a conjunct in $Cs$, and

2. there exists no total valuation $\sigma$ for the variables in $\mathcal{C}$ such that $S, \sigma \models_{cs} \tau = t$ and $S \models_{TR} C\sigma$, and

3. $A \in \mathcal{X}(S, t)$, where $\mathcal{X}(S, t)$ is as defined in Section 8.3.

# 9. Cycle Theory

The behaviour of KGP agents results from the application of transitions in sequences, repeatedly changing the state of the agent. These sequences are not fixed a priori, as in conventional agent architectures, but are determined dynamically by reasoning with declarative cycle theories, giving a form of flexible control. Cycle theories are given in the framework of Logic Programming with Priorities (LPP) as discussed in Section 3.

## 9.1 Formalisation of Cycle Theories.

Here we use the following new notations:

- $T(S, X, S', t)$ to represent the application of transition $T$ at time $t$ in state $S$ given input $X$ and resulting in state $S'$, and

- $*T(S, X)$ to represent that transition $T$ can potentially be chosen as the next transition in state $S$, with input $X$.

Recall that, for some of the transitions, $X$ may be the empty set $\{\}$, as indicated in Section 7. Formally, a cycle theory $\mathcal{T}_{cycle}$ consists of the following parts.

- An *initial* part $\mathcal{T}_{initial}$, that determines the possible transitions that the agent could perform when it starts to operate. Concretely, $\mathcal{T}_{initial}$ consists of rules of the form

    $*T(S_0, X) \leftarrow C(S_0, X)$

    which we refer to via the name $\mathcal{R}_{0|T}(S_0, X)$. These rules sanction that, if conditions $C$ hold in the initial state $S_0$ then the initial transition could be $T$, applied to state $S_0$ and input $X$. For example, the rule

    $\mathcal{R}_{0|GI}(S_0, \{\}) : *GI(S_0, \{\}) \leftarrow empty\_forest(S_0)$

    sanctions that the initial transition should be GI, if the forest in the initial state $S_0$ is empty.

    Note that $C(S_0, X)$ may be empty, and, if non-empty, $C(S_0, X)$ may refer to the current time via a condition $time\_now(t)$. For example, the rule

    $\mathcal{R}_{0|PI}(S_0, G) : *PI(S_0, G) \leftarrow Gs = f_{GS}(S_0, t), Gs \neq \{\}, G \in Gs, time\_now(t)$

    sanctions that the initial transition should be PI, if the forest in the initial state $S_0$ contains some goal that can be planned for at the current time (in that the goal selection operator picks that goal).

- A *basic* part $\mathcal{T}_{basic}$ that determines the possible transitions following given transitions, and consists of rules of the form

    $*T'(S', X') \leftarrow T(S, X, S', t), EC(S', X')$

329

which we refer to via the name $\mathcal{R}_{T|T'}(S', X')$. These rules sanction that, after transition $T$ has been executed, starting at time $t$ in the state $S$ and resulting in state $S'$, and the conditions $EC$ evaluated in $S'$ are satisfied, then transition $T'$ could be the next transition to be applied in $S'$, with input $X'$.[16] $EC$ are *enabling conditions* as they determine when $T'$ can be applied after $T$. They also determine input $X'$ for $T'$, via calls to selection operators. As for the initial part of $\mathcal{T}_{cycle}$, $EC$ may be empty and, if not, may refer to the current time. For example, the rule

$$\mathcal{R}_{AE|PI}(S', G) : *PI(S', G) \leftarrow \quad AE(S, As, S', t),$$
$$Gs = f_{GS}(S', t'), Gs \neq \{\}, G \in Gs, time\_now(t')$$

sanctions that PI should follow AE if at the current time there is some goal in the current state that is selected by the goal selection function.

- A *behaviour* part $\mathcal{T}_{behaviour}$ that contains rules describing dynamic priorities amongst rules in $\mathcal{T}_{basic}$ and $\mathcal{T}_{initial}$. Rules in $\mathcal{T}_{behaviour}$ are of the form

  $$\mathcal{R}_{T|T'}(S, X') \succ \mathcal{R}_{T|T''}(S, X'') \leftarrow BC(S, X', X'')$$

  with $T' \neq T''$, which we will refer to via the name $\mathcal{P}^T_{T' \succ T''}$. Recall that $\mathcal{R}_{T|T'}(\cdot)$ and $\mathcal{R}_{T|T''}(\cdot)$ are (names of) rules in $\mathcal{T}_{basic} \cup \mathcal{T}_{initial}$. Note that, with an abuse of notation, $T$ could be 0 in the case that one such rule is used to specify a priority over the *first* transition to take place, in other words, when the priority is over rules in $\mathcal{T}_{initial}$. These rules in $\mathcal{T}_{behaviour}$ sanction that, after transition $T$, if the conditions $BC$ hold, then we prefer the next transition to be $T'$ over $T''$. The conditions $BC$ are *behaviour conditions* as they give the behavioural profile of the agent. For example, the rule

  $$\mathcal{P}^T_{GI \succ T'} : \mathcal{R}_{T|GI}(S, \{\}) \succ \mathcal{R}_{T|T'}(S, X) \leftarrow empty\_forest(S)$$

  sanctions that GI should be preferred to any other transition after any transition that results into a state with an empty forest. As for the other components of $\mathcal{T}_{cycle}$, the conditions $BC$ may refer to the current time.

- An *auxiliary part* including definitions for any predicates occurring in the enabling and behaviour conditions.

- An *incompatibility part*, in effect expressing that only one (instance of a) transition can be chosen at any one time.

Hence, $\mathcal{T}_{cycle}$ is an LPP-theory where: (i) $P = \mathcal{T}_{initial} \cup \mathcal{T}_{basic}$, and (ii) $H = \mathcal{T}_{behaviour}$.

## 9.2 Operational Trace

The cycle theory $\mathcal{T}_{cycle}$ of an agent is responsible for its behaviour, in that it induces an *operational trace* of the agent, namely a (typically infinite) sequence of transitions

$$T_1(S_0, X_1, S_1, t_1), \ldots, T_i(S_{i-1}, X_i, S_i, t_i), T_{i+1}(S_i, X_{i+1}, S_{i+1}, t_{i+1}), \ldots$$

such that

---

16. Note that in order to determine that $T'$ is a possible transition after $T$, with a rule of the earlier form, one only needs to know that $T$ has been applied and resulted into the state $S'$. This is conveyed by the choice of name: $\mathcal{R}_{T|T'}(S', X')$. In other words, by using a Prolog notation, we could have represented the rule as $*T'(S', X') \leftarrow T(\_, \_, S', \_), EC(S', X')$. Thus, the rule is "Markovian".

- $S_0$ is the given initial state;

- for each $i \geq 1$, $t_i$ is given by the clock of the system $(t_i < t_{i+i})$;

- $(\mathcal{T}_{cycle} - \mathcal{T}_{basic}) \cup \{time\_now(t_1)\} \models_{pr} *T_1(S_0, X_1)$;

- for each $i \geq 1$

  $(\mathcal{T}_{cycle} - \mathcal{T}_{initial}) \cup \{T_i(S_{i-1}, X_i, S_i, t_i), time\_now(t_{i+1})\} \models_{pr} *T_{i+1}(S_i, X_{i+1})$

namely each (non-final) transition in a sequence is followed by the most preferred transition, as specified by $\mathcal{T}_{cycle}$. If, at some stage, the most preferred transition determined by $\models_{pr}$ is not unique, we choose one arbitrarily.

### 9.3 Normal Cycle Theory

The *normal cycle theory* is a concrete example of cycle theory, specifying a pattern of operation where the agent prefers to follow a sequence of transitions that allows it to achieve its goals in a way that matches an expected "normal" behaviour. Other examples of possible cycle theories can be found in the literature (Kakas, Mancarella, Sadri, Stathis, & Toni, 2005; Sadri & Toni, 2006).

Basically, the "normal" agent first introduces goals (if it has none to start with) via GI, then reacts to them, via RE, and then repeats the process of planning for them, via PI, executing (part of) the chosen plans, via AE, revising its state, via SR, until all goals are dealt with (successfully or revised away). At this point the agent returns to introducing new goals via GI and repeating the above process. Whenever in this process the agent is interrupted via a passive observation, via POI, it chooses to introduce new goals via GI, to take into account any changes in the environment. Whenever it has actions which are "unreliable", in the sense that their preconditions definitely need to be checked, the agent senses them (via SI) before executing the action. Whenever it has actions which are "unreliable", in the sense that their effects definitely need to be checked, the agent actively introduces actions that aim at sensing these effects, via AOI, after having executed the original actions. If initially the agent is equipped with some goals, then it would plan for them straightaway by PI.

The full definition of the normal cycle theory is given in the appendix. This is used to provide the control in the examples of the next section. Here, note that, although the normal cycle theory is based on the classic observe-plan-act cycle of agent control, it generalises this in several ways giving more flexibility on the agent behaviour to adapt to a changing environment. For example, the goals of the agent need not be fixed but can be dynamically changed depending on newly acquired information. Let us illustrates this feature with a brief example here. Suppose that the current state of our agent contains the top-level non-reactive goal $\langle return\_home(\tau_1), \{\tau_1 < 7\}\rangle$ and that a POI occurs which adds an observation $observed(low\_battery, 2)$ at time 2. A subsequent GI transition generated by the normal cycle theory introduces a new goal $\langle recharge\_battery(\tau_2), \{\tau_2 < 3\}\rangle$ which, depending on the details of $KB_{GD}$, either replaces the previous goal or adds this as an additional goal. The normal cycle theory will next choose to do a PI transition for the new and more urgent goal of recharging its battery.

## 10. Examples

In this section we revisit the examples introduced in Section 2.6 and used throughout the paper to illustrate the various components of the KGP model. Overall, the aim here is to illustrate the interplay of the transitions, and how this interplay provides the variety of behaviours afforded by the KGP model, including reaction to observations, generation and execution of conditional plans, and dynamic adjustment of goals and plans.

Unless specified differently, we will assume that $\mathcal{T}_{cycle}$ will be the *normal cycle theory* presented in Section 9.3. We will provide any domain-dependent definition in the auxiliary part of $\mathcal{T}_{cycle}$ explicitly, where required.

### 10.1 Setting 1 Formalised

We formalise here the initial state, knowledge bases and behaviour of *svs* for Setting 1 described in Section 2.6.1.

#### 10.1.1 Initial State

For simplicity, the observations, goals and the plan of *svs* can be assumed to be empty initially. More concretely let the (initial) state of *svs* be

$$
\begin{aligned}
KB_0 &= \{ \, \} \\
\mathcal{F} &= \{ \, \} \\
\mathcal{C} &= \{ \, \} \\
\Sigma &= \{ \, \}
\end{aligned}
$$

#### 10.1.2 Knowledge Bases

Following Section 5.1.4, we formulate the reactivity knowledge base for agent *svs* in terms of the utterances $query\_ref$, $refuse$, $inform$ inspired by the FIPA specifications for communicative acts (FIPA, 2001a, 2001b). However, although we use the same names of communicative acts as in the FIPA specification, we do not adopt here their "mentalistic" semantic interpretation in terms of pre- and post-conditions. Thus, $KB^{svs}_{react}$ is formulated as:

$observed(C, tell(C, svs, query\_ref(Q), D, T0), T), \; holds\_at(have\_info(Q, I), T)$
$\quad \Rightarrow assume\_happens(tell(svs, C, inform(Q, I), D), T'), \; T' > T$

$observed(C, tell(C, svs, query\_ref(Q), D, T0), T), \; holds\_at(no\_info(Q), T)$
$\quad \Rightarrow assume\_happens(tell(svs, C, refuse(Q), D), T'), \; T' > T$

$assume\_happens(tell(svs, C, inform(Q, I), D), T),$
$assume\_happens(tell(svs, C, refuse(Q), D), T')$
$\quad \Rightarrow false$

$assume\_happens(A, T), not\; executable(A) \Rightarrow false$

$executable(tell(svs, C, S, D)) \leftarrow C \neq svs$

$initially(no\_info(arrival(tr01))$

$precondition(tell(svs, C, inform(Q, I), D), have\_info(Q, I))$

$initiates(tell(C, svs, inform(Q, I), D), T, have\_info(Q, I))$

$terminates(tell(C, svs, inform(Q, I), D), T, no\_info(Q))$

### 10.1.3 Behaviour

To illustrate the behaviour of the *psa* we will assume that this agent requests from *svs*, at time 3, say, the arrival time of tr01. *svs* receives a request from *psa* at time 5 for the arrival time of $tr01$. Via POI at time 5 *svs* records in its $KB_0$:

$observed(psa, tell(psa, svs, query\_ref(arrival(tr01)), d, 3), 5)$

where $d$ is the dialogue identifier. Then, via RE, at time 7, say, *svs* modifies its state by adding to $\mathcal{F}$ a tree $\mathcal{T}$ rooted at an action $a_1$ to answer to *psa*. This action $a_1$ is a refusal represented as:

$a_1 = tell(svs, psa, refuse(arrival(tr01)), d, \tau),$

and the temporal constraint $\tau > 7$ is added to $\mathcal{C}$.

The refusal action is generated via the Reactivity capability because *svs* does not have information about the requested arrival time. *svs* executes the planned action $a_1$ at time 10, say, via the AE transition, instantiating its execution time, adding the following record to $KB_0$:

$executed(tell(svs, psa, refuse(arrival(tr01)), d), 10),$

and updating $\Sigma$ by adding $\tau = 10$ to it.

Suppose then that *svs* makes two observations then as follows. At time 17 *svs* receives information of the arrival time (18) of the $tr01$ train from *co*. Via POI, *svs* records in its $KB_0$ [17]:

$observed(co, tell(co, svs, inform(arrival(tr01), 18), d', 15), 17).$

Assume further that at time 25 *svs* receives another request from *psa* about the arrival time of $tr01$ and, via POI, *svs* records in its $KB_0$:

$observed(psa, tell(psa, svs, query\_ref(arrival(tr01)), d'', 20), 25)$

with a new dialogue identifier $d''$. This leads to a different answer from *svs* to the query of *psa*. *svs* adds an action to its state to answer *psa* with the arrival time. This is done again via RE, say at time 28. A new tree is added in $\mathcal{F}$ rooted at the (reactive) action

$tell(svs, psa, inform(arrival(tr01), 18), d'', \tau'),$

and the new temporal constraint $\tau' > 28$ is added to $\mathcal{C}$.

Via AE, *svs* executes the action, instantiating its execution time to 30, say, and adding the following record

---

17. $d'$ is the identifier of the dialogue within which this utterance has been performed, and would typically be different from the earlier $d$.

$$executed(tell(svs, psa, inform(arrival(tr01), 18), d''), 30)$$

to $KB_0$, and adding $\tau' = 30$ to $\Sigma$.

Eventually, SR will clear the planned (and executed) actions from the $\mathcal{F}$ component of the state of $svs$.

## 10.2 Setting 2 Formalised

We formalise here the initial state, knowledge bases and behaviour of $psa$ for Setting 2 described in Section 2.6.2.

### 10.2.1 Initial State

Let us assume that initially the state of $psa$ is as follows:

$$
\begin{aligned}
KB_0 &= \{ \} \\
\mathcal{F} &= \{\mathcal{T}_1, \mathcal{T}_2\} \\
\mathcal{C} &= \{\tau_1 < 15, \tau_2 < 15\} \\
\Sigma &= \{ \}
\end{aligned}
$$

where $\mathcal{T}_1$ and $\mathcal{T}_2$ consist of a goals (respectively):

$g_1 = have\_ticket(madrid, denver, \tau_1)$ and
$g_2 = have\_visa(usa, \tau_2)$.

### 10.2.2 Knowledge Bases

To plan for goal $g_1$, the $KB_{plan}^{psa}$ contains:

$initiates(buy\_ticket\_online(From, To), T, have\_ticket(From, To))$
$precondition(buy\_ticket\_online(From, To), available\_connection)$
$precondition(buy\_ticket\_online(From, To), available\_destination(To))$.

To plan for goal $g_2$, the $KB_{plan}^{psa}$ contains:

$initiates(apply\_visa(usa), T, have\_visa(usa))$
$precondition(apply\_visa(usa), have\_address(usa))$
$initiates(book\_hotel(L), T, have\_address(usa)) \leftarrow holds(in(L, usa), T)$.

### 10.2.3 Behaviour

When PI is called on the above state, at time 2, say, it generates a partial plan for the goal, changing the state as follows. The goal $g_1$ acquires three children in $\mathcal{T}_1$. These are:

$g_{11} = available\_connection(\tau_{11})$,
$g_{12} = available\_destination(denver, \tau_{12})$,
$a_{13} = buy\_ticket\_online(madrid, denver, \tau_{13})$.

Also, consequently, the set of temporal constraints is updated to:

$\mathcal{C} = \{\tau_1 < 15, \tau_2 < 15, \tau_{11} = \tau_{13}, \tau_{12} = \tau_{13}, \tau_{13} < \tau_1, \tau_1 > 2\}$.

The action $a_{13}$ is generated as an action that initiates goal $g_1$. Moreover, every plan that is generated must satisfy the integrity constraints in $KB_{plan}$. In particular, any precondition of actions in the tree that do not already hold must be generated as sub-goals in the tree. This is why $g_{11}$ and $g_{12}$ are generated in the tree as above.

Now via the transition SI, the following sensing actions are added to $\mathcal{T}_1$ as siblings of action $a_{13}$ [18]:

$a_{14} = sense(available\_connection, \tau_{14})$
$a_{15} = sense(available\_destination(denver), \tau_{15})$

and the constraints

$\tau_{14} = \tau_{15}, \tau_{14} < \tau_{13}$

are added to $\mathcal{C}$.

Then, via AE, these two sensing actions are executed (before the original action $a_1$), and $KB_0$ is updated with the result of the sensing as follows. Suppose these two actions are executed at time 5. Consider the first action that senses the fluent $available\_connection$. If this fluent is confirmed by the physical sensing capability, i.e. if $available\_connection : true$ is in X such that

$\texttt{sensing}(\{available\_connection, available\_destination\}, 5) = X,$

then $observed(available\_connection, 5)$ is added to $KB_0$. On the other hand, if

$$available\_connection : false$$

is in X as above, then $observed(\neg available\_connection, 5)$ is added to $KB_0$. In both cases $\tau_{14} = 5$ is added to $\Sigma$.

If neither of these cases occurs, i.e. if the sensing capability cannot confirm either of $available\_connection$ or $\neg available\_connection$, then no fact is added to $KB_0$. Similarly for the other precondition, $available\_destination$. Let us assume that after this step of AE, $KB_0$ becomes

$observed(available\_connection, 5)$
$observed(available\_destination(denver), 5)$

AE can then execute the original action $a_{13}$. Note that the agent might decide to execute the action even if one or both preconditions are not known to be satisfied after the sensing. If $g_1$ is achieved, SR will eliminate it and $a_{13}$, $a_{14}$, $a_{15}$, $g_{11}$, $g_{12}$ from the state. In the resulting state, $\mathcal{F} = \{\mathcal{T}_2\}$, and PI is called, say at time 6. This results in generating a partial plan for $g_2$, and changing the state so that in $\mathcal{T}_2$ the root $g_2$ has children

$a_{21} = apply\_visa(usa, \tau_{21})$
$g_{22} = have\_address(usa, \tau_{22})$

and $\tau_{21} < \tau_2, \tau_{22} = \tau_{21}$ are added to $\mathcal{C}$. Then, further PI, say at time 7, introduces

$a_{23} = book\_hotel(denver, \tau_{23})$

---

18. For this we assume that the auxiliary part of $\mathcal{T}_{cycle}$ contains the rule
    $unreliable\_pre(As) \leftarrow buy\_ticket\_online(\_, \_, \_) \in As$

as a child of $g_{22}$ in $\mathcal{T}_2$, and adding $\tau_{23} < \tau_{22}$ to $\mathcal{C}$. Then, AE at time 8 executes $a_{23}$, adding it to $KB_0$, and further AE at time 9 executes $a_{22}$, also updating $KB_0$. Finally, SR eliminates all actions and goals in $\mathcal{T}_2$ and returns an empty $\mathcal{F}$ in the state.

## 11. Related Work

Many proposals exist for models and architectures of individual agents based on computational logic foundations (see e.g. the survey by Fisher, Bordini, Hirsch, & Torroni, 2007). Some of these proposals are based on logic programming, for example IMPACT (Arisha, Ozcan, Ross, Subrahmanian, Eiter, & Kraus, 1999; Subrahmanian, Bonatti, Dix, Eiter, Kraus, Ozcan, & Ross, 2000), AAA (Balduccini & Gelfond, 2008; Baral & Gelfond, 2001), DALI (Costantini & Tocchio, 2004), MINERVA (Leite, Alferes, & Pereira, 2002), GOLOG (Levesque, Reiter, Lesperance, Lin, & Scherl, 1997), and IndiGolog (De Giacomo, Levesque, & Sardiña, 2001). Other proposals are based on modal logic or first-order logic approaches, for example the BDI model (Bratman et al., 1988; Rao & Georgeff, 1997) and its extensions to deal with normative reasoning (Broersen, Dastani, Hulstijn, Huang, & van der Torre, 2001), Agent0 (Shoham, 1993), AgentSpeak (Rao, 1996) and its variants, 3APL (Hindriks, de Boer, van der Hoek, & Meyer, 1999) and its variants (Dastani, Hobo, & Meyer, 2007).

At a high level of comparison there are similarities in the objectives of most existing computational logic models of agency and KGP, in that they all aim at specifying knowledge-rich agents with certain desirable behaviours. There are also some similarities in the finer details of the KGP model and some of the above related work, as well as differences.

A feature of the KGP which, to the best of our knowledge, is novel is the declarative and context-sensitive specification of an agent's cycle. To avoid a static cycle of control (Rao & Georgeff, 1991; Rao, 1996), KGP relies upon a *cycle theory* which determines, at run time, given the circumstances and the individual profile of the agent, what the next step should be. The cycle theory is sensitive to both solicited and unsolicited information that the agent receives from its environment, and helps the agent to adapt its behaviour to the changes it experiences. The approach closest to our work is that of 3APL (Hindriks et al., 1999) as extended by Dastani, de Boer, Dignum, and Meyer (2003), which provides meta-programming constructs for specifying the cycle of an agent such as goal selection, plan expansion, execution, as well as if-then-else and while-loop statements. Unlike the imperative constructs of 3APL, KGP uses a set of selection operators that can be extended to model different behaviours and types of agents. A flexible ordering of transitions is then obtained using preference reasoning about which transitions can be applied at a specific point in time. These preferences may change according to external events or changes in the knowledge of the agent.

Another central distinguishing feature of the KGP model, in comparison with existing models, including those based on logic programming, is its modular integration within a single framework of abductive logic programming, temporal reasoning, constraint logic programming, and preference reasoning based on logic programming with priorities, in order to support a diverse collection of capabilities. Each one of these is specified declaratively and equipped with its own provably correct computational counterpart (see Bracciali,

Demetriou, Endriss, Kakas, Lu, Mancarella, Sadri, Stathis, Terreni, & Toni, 2004, for a detailed discussion).

Compared with existing logic programming approaches KGP has two main similarities with MINERVA (Leite et al., 2002), an architecture that exploits computational logic and gives both declarative and operational semantics to its agents. Unlike KGP, a MINERVA agent consists of several specialised, possibly concurrent, sub-agents performing various tasks, and relies upon MDLP (Multidimensional Dynamic Logic Programming) (Leite et al., 2002). MDLP is the basic knowledge representation mechanism of an agent in MINERVA, which is based on an extension of answer-set programming and explicit rules for updating the agent's knowledge base. In KGP instead we integrate abductive logic programming and logic programming with priorities combined with temporal reasoning.

Closely related to our work in KGP is the logic-based agent architecture for reasoning agents of Baral and Gelfond (2001). This architecture assumes that the state of an agent's environment is described by a set of fluents that evolve over time in terms of transitions labelled by actions. An agent is also assumed to be capable of correctly observing the state of the environment, performing actions, and remembering the history of what happened in it. The agent's knowledge base consists of an action description part specifying the internal agent transitions, which are domain specific and not generic as in KGP. The knowledge base also contains what the agent observes in the environment including its own actions, as in KGP's $KB_0$. The temporal aspects of agent transitions are specified in the action language $\mathcal{AL}$ implemented in *A-Prolog*, a language of logic programs under the answer-set programming semantics. The answer sets of domain specific programs specified in $\mathcal{AL}$ correspond to plans that in KGP are hypothetical narratives of the abductive event calculus. The control of the agent is based on a static *observe-think-act* cycle, an instance of the KGP cycle theories. A more recent and refined account of the overall approach has given rise to the AAA Architecture, see (Balduccini & Gelfond, 2008) for an overview.

DALI (Costantini & Tocchio, 2004) is a logic programming language designed for executable specification of logical agents. Like KGP, DALI attempts to provide constructs to represent reactivity and proactivity in an agent using extended logic programs. A DALI agent contains reactive rules, events, and actions aimed at interacting with an external environment. Behaviour (in terms of reactivity or proactivity) of a DALI agent is triggered by different event types: external, internal, present, and past events. All the events and actions are time stamped so as to record when they occur. External events are like the observations in KGP, while past events are like past observations. However, KGP does not support internal events but has instead the idea of transitions that are called by the cycle theory to trigger reactive or proactive behaviour.

IndiGolog (De Giacomo et al., 2001) is a high-level programming language for robots and intelligent agents that supports, like KGP, on-line planning, sensing and plan execution in dynamic and incompletely known environments. It is a member of the Golog family of languages (Levesque et al., 1997) that use a Situation Calculus theory of action to perform the reasoning required in executing the program. Instead in the KGP model we rely on abductive logic programming and logic programming with priorities combined with temporal reasoning. Instead of the Situation Calculus in KGP we use the Event Calculus for temporal reasoning, but our use of the Event Calculus is not a prerequisite of the model as in InterRaP (Müller, Fischer, & Pischel, 1998), but can be replaced with another temporal reasoning framework, if needed. Apart from the difference between the use of the Situation and Event Calculi, in IndiGolog goals cannot be decided dynamically, whereas in the KGP model they change dynamically according to the specifications in the Goal Decision capability.

There is an obvious similarity of the KGP model with the BDI model (Bratman et al., 1988) given by the correspondence between KGP's knowledge, goals and plan and BDI's beliefs, desires and intentions, respectively. Apart from the fact that the BDI model is based on modal logic, in KGP the knowledge (beliefs in BDI) is partitioned in modules, to support the various reasoning capabilities. KGP also tries to bridge the gap between specification and the practical implementation of an agent. This gap has been criticized in BDI by Rao (1996), when he developed the AgentSpeak(L) language. The computational model of AgentSpeak(L) has been formally studied by d'Inverno and Luck (1998), while recent implementations of the AgentSpeak interpreter have been incorporated in the Jason platform (Bordini & Hübner, 2005). Like the KGP implementation in PROSOCS (Bracciali et al., 2006), the Jason implementation too seeks to narrow the gap between specification and executable BDI agent programs. Jason also extends BDI with new features like belief revision (Alechina, Bordini, Hübner, Jago, & Logan, 2006).

A particular line of work in BDI is that of Padgham and Lambrix (2005), who investigate how the notion of capability can be integrated in the BDI Logic of Rao and Georgeff (1991), so that a BDI agent can reason about its own capabilities. A capability in this work is informally understood as the ability to act rationally towards achieving a particular goal, in the sense of having an abstract plan type that is believed to achieve the goal. Formally, the BDI logic of Rao and Georgeff is extended to incorporate a modality for capabilities that constrains agent goals and intentions to be compatible with what the agent believes are its capabilities. A set of compatibility axioms are then presented detailing the semantic conditions to capture the desired inter-relationships among an agent's beliefs, capabilities, goals, and intentions. The work also summarises how the extensions of the BDI model can be implemented by adapting the BDI interpreter to include capabilities, further arguing the benefits of the extension over the original BDI Interpreter of Rao and Georgeff (1992).

In KGP capabilities equate to the reasoning capabilities of an agent that allow the agent to plan actions from a given state, react to incoming observations, or decide upon which goals to adopt. However, in KGP, we do not use capabilities at the level of an agent's domain specific knowledge to guide the agent in determining whether or not it is rational to adopt a particular goal.

The issue of the separation between specification and implementation exists between the KGP model and Agent0 (Shoham, 1993), and its later refinement PLACA (Thomas, 1995). Two other differences between the KGP and Agent0 and PLACA are the explicit links that exist in the KGP model amongst the goals (in the structuring of the forest in the agent state) and the richer theories in the KGP that specify priorities amongst potential goals which are not restricted to temporal orderings. These explicit links are exploited when revising goals and state, via the Revision transition, in the light of new information or because of the passage of time.

The BOID architecture (Broersen et al., 2001) extends the well known BDI model (Rao & Georgeff, 1992) with obligations, thus giving rise to four main components in representing an agent: beliefs, obligations, intentions and desires. The focus of BOID is to find ways of resolving conflicts amongst these components. In order to do so they define agent types, including some well known types in agent theories such as realistic, selfish, social and simple minded agents. The agent types differ in that they give different priorities to the rules for each of the four components. For instance, the simple minded agent gives higher priority to intentions, compared to desires and obligations, whereas a social agent gives higher priority to obligations than desires. They use priorities with propositional logic formulae to specify the four components and the agent types.

The existing KGP model already resolves some of the conflicts that BOID tries to address. For example, if there is a conflict between a belief and a prior intention, which means that an intended action can no longer be executed due to the changes in the environment, the KGP agent will notice this and will give higher priority to the belief than the prior intention, allowing the agent in effect to retract the intended action and, time permitting, to replan for its goals. The KGP model also includes a notion of priority used in the Goal Decision capability and the cycle theory that controls the behaviour of the agent. The KGP model has also been extended to deal with normative concepts, the extended model is known as N-KGP (Sadri, Stathis, & Toni, 2006). What N-KGP has in common with BOID is that it seeks to extend KGP with the addition of obligations. The N-KGP model also extends the notion of priorities by incorporating them amongst different types of goals and actions. A detailed comparison of N-KGP with related work is presented by Sadri, Stathis, and Toni (2006).

There are features that are included in some other approaches that are absent in the KGP model. BDI and, more so, the IMPACT system (Arisha et al., 1999; Subrahmanian et al., 2000) allow agents to have in their knowledge bases representations of the knowledge of other agents. These systems allow the agents both some degree of introspection and the ability to reason about other agents' beliefs and reasoning. The KGP model to this date does not include any such features. IMPACT also allows the incorporation of legacy systems, possibly using diverse languages, and has a richer knowledge base language including deontic concepts and probabilities. Similarly, the 3APL, system is based on a combination of imperative and logic programming languages, and includes an optimisation component absent from the KGP. This component in 3APL includes rules that identify if in a given situation the agent is pursuing a suboptimal plan, and help the agent find a better way of achieving its goals. 3APL also includes additional functionalities such as learning (van Otterlo, Wiering, Dastani, & Meyer, 2003), which our model does not currently support. 2APL (Dastani et al., 2007) is an extension of 3APL with goals and goal-plan rules as well as external and internal events. 2APL has a customisable (via graphical interface) cycle which is fixed once customised.

## 12. Conclusions

We have presented the computational logic foundations of the KGP model of agency. The model allows the specification of heterogeneous agents that can interact with each other, and can exhibit both proactive and reactive behaviour allowing them to function in dynamic environments by adjusting their goals and plans when changes happen in such environments. KGP incorporates a highly modular agent architecture that integrates a collection of reasoning and sensing capabilities, synthesised within transitions, orchestrated by cycle theories that take into account the dynamic context and agent preferences.

The formal specification of the KGP components within computational logic has the major advantage of facilitating both a formal analysis of the model and a direct verifiable implementation. This formal analysis has been started by Sadri and Toni (2006), where we give a formal analysis of KGP agents by exploring their effectiveness in terms of *goal achievement*, and *reactive awareness*, and the impact of their reasoning capabilities towards *progress* in goal achievement. An implementation of a precursor of this model, described by Kakas et al. (2004b), has already been developed within the PROSOCS platform of Stathis et al. (2004) upon provably correct computational counterparts defined for each component of the model as given by Kakas et al. (2004b). Concrete choices for these computational counterparts have been described by Bracciali et al. (2004). The resulting development framework allows the deployment and testing of the functionality of the earlier variant of KGP agents. Deployment of these agents relies upon the *agent template* designed by Stathis et al. (2002), which builds upon previous work with the *head/body* metaphor described by Steiner et al. (1991) and Haugeneder et al. (1994), and the *mind/body* architecture introduced by Bell (1995) and recently used by Huang, Eliens, and de Bra (2001). This development platform has been applied to a number of practical applications, and, in particular, to ambient intelligence by Stathis and Toni (2004). Also, Sadri (2005) has provided guidelines for specifying applications using KGP agents. Future work includes implementing and deploying the revised KGP model given in this paper: we envisage that this will pose limited conceptual challenges, as we will be able to capitalise on our experience in implementing and deploying the precursor of this model.

Sadri, Stathis, and Toni (2006) have explored how the precursor of the KGP agent model can be augmented with normative features allowing agents to reason about and choose between their *social* and *personal* goals, prohibitions and obligations. It would be interesting to continue this work for the finalised KGP model given in this paper.

Sadri and Toni (2005) have developed a number of different profiles of behaviour, defined in terms of specific cycle theories, and formally proved their advantages in given circumstances. It would be interesting to explore this dimension further, to characterise different agent personalities and provide guidance, through formal properties, as to the type of personality needed for applications.

Future work also includes extending the model to incorporate (i) other reasoning capabilities, including knowledge revision (e.g. by Inductive Logic Programming), and more sophisticated forms of temporal reasoning, including identifying explanations for unexpected observations, (ii) introspective reasoning and reasoning about the beliefs of other agents, (iii) further experimentation with the model via its implementation, and (iv) development of a concurrent implementation.

## Acknowledgments

This work was supported by the EU FET Global Computing Initiative, within the SOCS project (IST-2001-32530). We wish to thank all our colleagues in SOCS for useful discussions during the development of KGP. We are also grateful to Chitta Baral and the anonymous referees for helpful comments on an earlier version of this paper.

## Appendix A. Normal Cycle Theory

We give here the main parts of the normal $\mathcal{T}_{cycle}$, but exclude others, for example the definitions for *incompatible* and the auxiliary part, including definitions for predicates such as *empty_forest*, *unreliable_pre* etc. For more details see (Kakas et al., 2005).

$\mathcal{T}_{initial}$: This consists of the following rules:

$\quad \mathcal{R}_{0|GI}(S_0, \{\}) : *GI(S_0, \{\}) \leftarrow empty\_forest(S_0)$

$\quad \mathcal{R}_{0|AE}(S_0, As) : *AE(S_0, As) \leftarrow empty\_non\_executable\_goals(S_0), As = f_{AS}(S_0, t),$
$\qquad\qquad As \neq \{\}, time\_now(t)$

$\quad \mathcal{R}_{0|PI}(S_0, G) : *PI(S_0, G) \leftarrow Gs = f_{GS}(S_0, t), Gs \neq \{\}, G \in Gs, time\_now(t)$

$\mathcal{T}_{basic}$: This consists of the following rules:

- The rules for deciding what might follow an AE transition are as follows:

$\quad \mathcal{R}_{AE|PI}(S', G) : *PI(S', G) \leftarrow AE(S, As, S', t), Gs = f_{GS}(S', t'), Gs \neq \{\},$
$\qquad\qquad G \in Gs, time\_now(t')$

$\quad \mathcal{R}_{AE|AE}(S', As') : *AE(S', As') \leftarrow AE(S, As, S', t), As' = f_{AS}(S', t'),$
$\qquad\qquad As' \neq \{\}, time\_now(t')$

$\quad \mathcal{R}_{AE|AOI}(S', Fs) : *AOI(S', Fs) \leftarrow AE(S, As, S', t), Fs = f_{ES}(S', t'),$
$\qquad\qquad Fs \neq \{\}, time\_now(t')$

$\quad \mathcal{R}_{AE|SR}(S') : *SR(S', \{\}) \leftarrow AE(S, As, S', t)$

$\quad \mathcal{R}_{AE|GI}(S', \{\}) : *GI(S', \{\}) \leftarrow AE(S, As, S', t)$

Namely, AE could be followed by another AE, or by a PI, or by an AOI, or by a SR, or by a GI, or by a POI.

- The rules for deciding what might follow SR are as follows

$\quad \mathcal{R}_{SR|PI}(S', G) : *PI(S', G) \leftarrow SR(S, \{\}, S', t), Gs = f_{GS}(S', t'), Gs \neq \{\}, G \in Gs,$
$\qquad\qquad time\_now(t')$

$\quad \mathcal{R}_{SR|GI}(S', \{\}) : *GI(S', \{\}) \leftarrow SR(S, \{\}, S', t), Gs = f_{GS}(S', t'), Gs = \{\},$
$\qquad\qquad time\_now(t')$

$\quad \mathcal{R}_{SR|AE}(S', As) : *AE(S', As) \leftarrow SR(S, \{\}, S', t), As = f_{GS}(S', t'), As \neq \{\},$
$\qquad\qquad time\_now(t')$

Namely, SR can only be followed by PI or GI or AE, depending on whether or not there are goals to plan for in the state.

- The rules for deciding what might follow PI are as follows

$\quad \mathcal{R}_{PI|AE}(S', As) : *AE(S', As) \leftarrow PI(S, G, S', t), As = f_{AS}(S', t'), As \neq \{\},$
$\qquad\qquad time\_now(t')$

$\quad \mathcal{R}_{PI|SI}(S', Ps) : *SI(S', Ps) \leftarrow PI(S, G, S', t), Ps = f_{PS}(S', t'), Ps \neq \{\}, time\_now(t')$

The second rule is here to allow the possibility of sensing the preconditions of an action before its execution.

- The rules for deciding what might follow GI are as follows

$\quad \mathcal{R}_{GI|RE}(S', \{\}) : *RE(S', \{\}) \leftarrow GI(S, \{\}, S', t)$

$\quad \mathcal{R}_{GI|PI}(S', G) : *PI(S', G) \leftarrow GI(S, \{\}, S', t), Gs = f_{GS}(S', t'), Gs \neq \{\}, G \in Gs,$
$\qquad\qquad time\_now(t')$

Namely, GI can only be followed by RE or PI, if there are goals to plan for.

- The rules for deciding what might follow RE are as follows

$$\mathcal{R}_{RE|PI}(S', G) : *PI(S', G) \leftarrow RE(S, \{\}, S', t), Gs = f_{GS}(S', t'), Gs \neq \{\}, G \in Gs,$$
$$time\_now(t')$$
$$\mathcal{R}_{RE|SI}(S', Ps) : *SI(S', Ps) \leftarrow RE(S, \{\}, S', t), Ps = f_{PS}(S', t'), Ps \neq \{\},$$
$$time\_now(t')$$

- The rules for deciding what might follow SI are as follows
$$\mathcal{R}_{SI|AE}(S', As) : *AE(S', As) \leftarrow SI(S, Ps, S', t), As = f_{AS}(S', t'), As \neq \{\},$$
$$time\_now(t')$$
$$\mathcal{R}_{SI|SR}(S', \{\}) : *SR(S', \{\}) \leftarrow SI(S, Ps, S', t)$$

- The rules for deciding what might follow AOI are as follows
$$\mathcal{R}_{AOI|AE}(S', As) : *AE(S', As) \leftarrow AOI(S, Fs, S', t), As = f_{AS}(S', t'), As \neq \{\},$$
$$time\_now(t')$$
$$\mathcal{R}_{AOI|SR}(S', \{\}) : *SR(S', \{\}) \leftarrow AOI(S, Fs, S', t)$$
$$\mathcal{R}_{AOI|SI}(S', Ps) : *SI(S', Ps) \leftarrow AOI(S, Fs, S', t), Ps = f_{PS}(S', t'), Ps \neq \{\},$$
$$time\_now(t')$$

- The rules for deciding what might follow POI are as follows
$$\mathcal{R}_{POI|GI}(S', \{\}) : *GI(S', \{\}) \leftarrow POI(S, \{\}, S', t)$$

$\mathcal{T}_{behaviour}$: This consists of the following rules:

- GI should be given higher priority if there are no trees in the state:
$$\mathcal{P}^T_{GI \succ T'} : \mathcal{R}_{T|GI}(S, \{\}) \succ \mathcal{R}_{T|T'}(S, X) \leftarrow empty\_forest(S)$$
for all transitions $T, T'$, $T' \neq GI$, and with $T$ possibly 0 (indicating that if there are no trees in the initial state of an agent, then GI should be its first transition).

- GI is also given higher priority after a POI:
$$\mathcal{P}^{POI}_{GI \succ T} : \mathcal{R}_{POI|GI}(S', \{\}) \succ \mathcal{R}_{POI|T}(S, \{\}, S')$$
for all transitions $T \neq GI$.

- After GI, the transition RE should be given higher priority:
$$\mathcal{P}^{GI}_{RE \succ T} : \mathcal{R}_{GI|RE}(S, \{\}) \succ \mathcal{R}_{GI|T}(S, X)$$
for all transitions $T \neq RE$.

- After RE, the transition PI should be given higher priority:
$$\mathcal{P}^{RE}_{PI \succ T} : \mathcal{R}_{RE|PI}(S, G) \succ \mathcal{R}_{RE|T}(S, X)$$
for all transitions $T \neq PI$.

- After PI, the transition AE should be given higher priority, unless there are actions in the actions selected for execution whose preconditions are "unreliable" and need checking, in which case SI will be given higher priority:
$$\mathcal{P}^{PI}_{AE \succ T} : \mathcal{R}_{PI|AE}(S, As) \succ \mathcal{R}_{PI|T}(S, X) \leftarrow not\ unreliable\_pre(As)$$
for all transitions $T \neq AE$.
$$\mathcal{P}^{PI}_{SI \succ AE} : \mathcal{R}_{PI|SI}(S, Ps) \succ \mathcal{R}_{PI|AE}(S, As) \leftarrow unreliable\_pre(As)$$

- After SI, the transition AE should be given higher priority
$$\mathcal{P}^{SI}_{AE \succ T} : \mathcal{R}_{SI|AE}(S, As) \succ \mathcal{R}_{SI|T}(S, X)$$
for all transitions $T \neq AE$.

- After AE, the transition AE should be given higher priority until there are no more actions to execute in the state, in which case either AOI or SR should be given higher priority, depending on whether there are actions which are "unreliable", in the sense that their effects need checking, or not:

$$\mathcal{P}^{AE}_{AE \succ T} : \mathcal{R}_{AE|AE}(S, As) \succ \mathcal{R}_{AE|T}(S, X)$$

for all transitions $T \neq AE$. Note that, by definition of $\mathcal{T}_{basic}$, the transition AE is applicable only if there are still actions to be executed in the state.

$$\mathcal{P}^{AE}_{AOI \succ T} : \mathcal{R}_{AE|AOI}(S, Fs) \succ \mathcal{R}_{AE|T}(S, X)) \leftarrow BC^{AE}_{AOI|T}(S, Fs, t), time\_now(t)$$

for all transitions $T \neq AOI$, where the behaviour condition $BC^{AE}_{AOI|T}(S, Fs, t)$ is defined (in the auxiliary part) by:

$$BC^{AE}_{AOI|T}(S, FS, t) \leftarrow empty\_executable\_goals(S, t), unreliable\_effect(S, t)$$

Similarly, we have:

$$\mathcal{P}^{AE}_{SR \succ T} : \mathcal{R}_{AE|SR}(S, \{\}) \succ \mathcal{R}_{AE|T}(S, X)) \leftarrow BC^{AE}_{SR|T}(S, t), time\_now(t)$$

for all transitions $T \neq SR$ where:

$$BC^{AE}_{SR|T}(S, t) \leftarrow empty\_executable\_goals(S, t), not\ unreliable\_effect(S, t)$$

Here, we assume that the auxiliary part of $\mathcal{T}_{cycle}$ specifies whether a given set of actions contains any "unreliable" action, in the sense expressed by $unreliable\_effect$, and defines the predicate $empty\_executable\_goals$.

- After SR, the transition PI should have higher priority:

$$\mathcal{P}^{SR}_{PI \succ T} : \mathcal{R}_{SR|PI}(S, G) \succ \mathcal{R}_{SR|T}(S, X))$$

for all transitions $T \neq PI$.

Note that, by definition of $\mathcal{T}_{basic}$, the transition PI is applicable only if there are still goals to plan for in the state. If there are no actions and goals left in the state, then rule $\mathcal{R}_{GI|T}$ would apply.

- In the initial state PI should be given higher priority:

$$\mathcal{P}^{0}_{PI \succ T} : \mathcal{R}_{0|PI}(S, G) \succ \mathcal{R}_{0|T}(S, X)$$

for all transitions $T \neq PI$. Note that, by definition of $\mathcal{T}_{initial}$ below, the transition PI is applicable initially only if there are goals to plan for in the initial state.